\pdfoutput=1

\documentclass[11pt]{article}
\usepackage[T2A]{fontenc} 
\usepackage[utf8]{inputenc} 
\usepackage[russian,english]{babel} 
\usepackage{xcolor}
\usepackage{amsmath}
\usepackage{acl}
\usepackage{threeparttable}
\usepackage{times}
\usepackage{latexsym}
\usepackage{pdfpages}
\usepackage[T1]{fontenc}
\usepackage{multirow}

\usepackage[utf8]{inputenc}
\makeatletter
\renewcommand\@makefntext[1]{\leftskip=0em\hskip1em\@makefnmark#1}
\makeatother
\usepackage{microtype}
\usepackage{booktabs}
\usepackage{makecell}
\usepackage{tipa}
\title{Enhancing Multilingual Capabilities of Large Language Models through Self-Distillation from Resource-Rich Languages}

\author{
Yuanchi Zhang$^1$, Yile Wang$^2$, Zijun Liu$^1$, Shuo Wang$^{*1}$, Xiaolong Wang$^1$, \\{\bf Peng Li}\thanks{ \quad Corresponding authors.}$^{\ \,2}${\bf , Maosong Sun}$^1${\bf , Yang Liu}$^{1,2}$ \\
$^1$ Department of Computer Science and Technology, Tsinghua University, Beijing, China \\
$^2$ Institute for AI Industry Research (AIR), Tsinghua University, Beijing, China \\
\texttt{yuanchi-21@mails.tsinghua.edu.cn;  
 wangyile@air.tsinghua.edu.cn}\\\texttt{ liuzijun20@mails.tsinghua.edu.cn; wangshuo.thu@gmail.com} \\
  \texttt{lipeng@air.tsinghua.edu.cn; \{sms,liuyang2011\}@tsinghua.edu.cn}}

\begin{document}
\maketitle
\begin{abstract}
While large language models (LLMs) have been pre-trained on multilingual corpora, their performance still lags behind in most languages compared to a few resource-rich languages. One common approach to mitigate this issue is to translate training data from resource-rich languages into other languages and then continue training. However, using the data obtained solely relying on translation while ignoring the original capabilities of LLMs across languages is not always effective, which we show will limit the performance of cross-lingual knowledge transfer. In this work, we propose SDRRL, a method based on \underline{S}elf-\underline{D}istillation from \underline{R}esource-\underline{R}ich \underline{L}anguages that effectively improve multilingual performance by leveraging the internal capabilities of LLMs on resource-rich languages. We evaluate on different LLMs (LLaMA-2 and SeaLLM) and source languages (English and French) across various comprehension and generation tasks, experimental results demonstrate that SDRRL can significantly enhance multilingual capabilities while minimizing the impact on original performance in resource-rich languages.\footnote{The source code is available at \url{https://github.com/THUNLP-MT/SDRRL}. }


\end{abstract}

\section{Introduction}

Contemporary large language models (LLMs; \citealp{openaichatgpt,openaigpt4,touvron2023llama,touvron2023llama2,jiang2023mistral,team2023gemini}) are predominantly trained on multilingual corpora. However, the language distribution in the data is highly imbalanced. For instance, LLMs like LLaMA-2~\cite{touvron2023llama2}, with English as the primary language, have also been trained on Japanese text, yet the quantity of English tokens used during pre-training exceeds that of Japanese by a factor of 897. 

\begin{figure}[t!]
    \centering
    \includegraphics[width=\columnwidth]{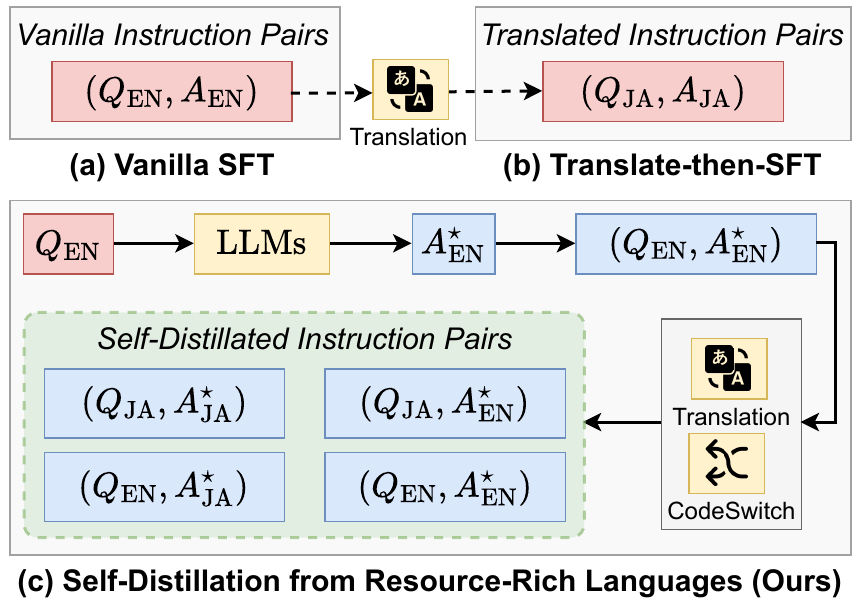}
    \caption{Comparison between vanilla supervised fine-tuning (SFT), translate-then-SFT, and our proposed method. Besides using the translated question-answer pairs in the target language (e.g., Japanese), SDRRL further leverages the generated answer $A^{\star}_{\rm EN}$ by LLMs in the resource-rich language (e.g., English) and collects self-distillated data (in green box) to help enhance its multilingual capabilities.}
    \label{fig:intro}
\end{figure}
The imbalanced data distribution above has led to significant limitations in the capabilities of LLMs across most languages. To enhance the multilingual capabilities, a common approach follows the translating and then supervised fine-tuning (SFT;~\citealp{ouyang2022training}) paradigm, as shown in Figure~\ref{fig:intro}(b). Specifically, training data is translated into the target language using either the model itself or an external machine translation (MT) system before continuing the training process, thereby offering more data in the target language and improving multilingual capabilities.

However, the translate-then-SFT method encounters several challenges: First, the multilingual enhancement gained from translated ``question-answer'' pairs is limited and may sometimes even degrade the capabilities in the original primary language~\cite{28zhu2024question}. Second, constrained by the accuracy of machine translation (especially for the low-resource languages), the translated texts used for training can be highly noisy, containing numerous awkward sentences and incorrect content, adversely affecting the quality of the generated text and the multilingual abilities of the LLMs. Therefore, we explore a new question along this trajectory: \textit{Besides translating the training pairs, can we enhance the abilities in other languages by leveraging the original relatively strong capabilities of LLMs in resource-rich language?}

In this paper, we introduce SDRRL, a method that uses \underline{S}elf-\underline{D}istillation from \underline{R}esource-\underline{R}ich \underline{L}anguages) to achieve the goal mentioned above. Specifically, as illustrated in Figure~\ref{fig:intro}(c), SDRRL comprises two parts: (1) \textit{Self-Distillation}: Instead of the ground-truth answer, responses from LLMs in resource-rich languages are collected to construct a transfer set. These are then translated into other languages using machine translation systems and code-switching tools, forming ``question-answer'' pairs that are semantically identical but linguistically varied, and conducting sentence-level knowledge self-distillation within the same batch. (2) \textit{Incorporating External Parallel Corpus}: We further involve a small amount of machine translation data in the distillation, aiming to align the linguistic representation spaces better and mitigate the negative impact of the noise in machine translation systems on the generative capabilities of LLMs.

Our experiments, based on LLaMA-2-7B~\cite{touvron2023llama2} and SeaLLM-7B~\cite{nguyen2023seallms} with English as the resource-rich language, demonstrate that even with a smaller set of English instruction data as the transfer set, SDRRL can effectively distill English capabilities into 14 other languages, showing effectiveness in both multilingual comprehension and generation tasks. Further analysis indicates that SDRRL helps preserve the original capabilities in high-resource languages and improves the quality of generated responses.


\section{Related Work}
\textbf{Multilingual Language Models.} Using multilingual data during the pre-training is a common approach to enhance the multilingual capabilities of LLMs~\citep{1li2022pretrained,2lample2019crosslingual,3workshop2023bloom,4lin2022fewshot,5xue2021mt5}. Despite being pre-trained and fine-tuned targeting a few resource-rich languages, recent instruction-following LLMs~\citep{touvron2023llama2,jiang2023mistral,7wang2023openchat} have been found to still possess significant multilingual understanding and generation capabilities~\cite{9bandarkar2023belebele,10Niklaus_2023}. However, limited by the imbalanced training data distribution~\citep{15yang2023bigtranslate}, the multilingual capabilities of these popular LLMs lag behind those of languages with abundant resources~\citep{16Pahune_2023}.


\noindent\textbf{Cross-Lingual Transfer.} To enhance the capabilities in languages with scarce resources, one line of work is cross-lingual transfer, where skills learned from one source language can be readily transferred to other languages~\citep{17etxaniz2023multilingual,18huang2023languages,19ranaldi2023empowering}. This has been approached by designing prompts that leverage LLMs to self-translate questions into resource-rich languages~\citep{33qin2023crosslingual}, or by utilizing external machine translation systems for assistance ~\citep{21zhao2024llama}. Efforts have also been made to distill synthetic data from high-resource languages to low-resource ones~\citep{22chai2024xcot}.
~\citet{25shaham2024multilingual} and~\citet{34kew2023turning} leverage similarities between languages to stimulate capabilities in others. Compared to their work, we focus on proficiency in the resource-rich language and leverage it to improve performance in other languages.

\noindent\textbf{Cross-Lingual Alignment.} Another line of work is cross-lingual alignment~\citep{26schuster2019crosslingual}. Given the scarcity of multilingual data, the construction of alignment data or loss functions of varying granularity can align mid- and low-resource languages with those that are resource-rich. This includes the construction of pre-training tasks using multilingual aligned lexicons~\citep{27chi2021improving}, alignment of word embeddings~\citep{31Wen_Yi_2023,36schuster2019crosslingual}, using aligned data on one side of a problem to improve mathematical reasoning processes~\citep{28zhu2024question}, and encouraging language switching in chain-of-thought (CoT;~\citealp{wei2022chain}) reasoning~\citep{22chai2024xcot}. \citet{30mao2024tuning} have leveraged the LLM's own capabilities to generate aligned data, while others have constructed it with the aid of external systems ~\citep{32ranaldi2023does,35chen2023breaking}. Deriving and constructing multilingual supervision signals from existing datasets overlooks the fact that the model's own responses in high-resource languages can also serve as effective supervision signals. We show in our experiments that self-distillation not only improves the LLM's multilingual performance but also helps maintain the performance in the original resource-rich languages.

\noindent\textbf{Knowledge Distillation.} Knowledge distillation~\citep{37hinton2015distilling} is a widely used method for transferring knowledge~\citep{45gou2021knowledge}. In the text generation domain, sequence-level knowledge distillation~\citep{38kim2016sequencelevel} has been used as a means of data augmentation in areas such as machine translation~\citep{39gordon2019explaining}.
In particular, \textit{self-distillation}~\citep{41zhang2019teacher,42,43pham2022revisiting} is often utilized to distill knowledge from one component of a model to another~\citep{47}, or from one stage of a model to another~\citep{46Yang_2019_CVPR}. In this work, we apply distilling knowledge between the different linguistic representation spaces within the same LLM to enhance multilingual capabilities.


\section{Method}

In this section, we first revisit the supervised fine-tuning (SFT) and translate-then-SFT paradigm, subsequently dividing the discussions into two parts of our proposed SDRRL. In the first part, we construct a transfer set using responses in the resource-rich language from LLMs through sentence-level self-distillation. In the second part, we employ parallel translation-based instruction data to further improve multilingual generation capabilities.

\subsection{SFT and Translate-then-SFT Paradigm}
We consider the given instruction dataset comprised of $N$ entries $\mathcal{D} = \{(\mathbf{x}_i, \mathbf{y}_i)\}_{i=1}^N$, where $\mathbf{x}_i$ symbolizes the input sentence (question) for the $i$-th data point, and $\mathbf{y}_i$ signifies the corresponding ground-truth response (answer).

\noindent\textbf{Supervised Fine-Tuning.} For a LLM $\mathcal{M}_{\theta}$ parameterized by a set of parameters $\theta$, which produces a response denoted as $\mathbf{\hat{y}} = \mathcal{M}_{\theta}(\mathbf{x})$ for the given input question $\mathbf{x}$, the objective of SFT is to align the output sentence $\mathbf{\hat{y}}$ as closely as possible with the ground-truth response $\mathbf{y}$. Specifically, the cross-entropy (CE) loss is employed to assess the discrepancy between the model output $\mathbf{\hat{y}}$ and the ground-truth output $\mathbf{y}$ for a single sample $(\mathbf{x}, \mathbf{y})$, defined as:
\begin{equation}
\ell_{\rm CE}(\mathbf{y}, \mathbf{\hat{y}}) = -\sum_{j=1}^{|\mathcal{V}|} y_j \log(\hat{y}_j)
\label{eq:sft}
\end{equation}
where $y_j$ is the one-hot encoding of the ground truth output $\mathbf{y}$ at position $j$, $\hat{y}_j$ is the probability of the model output $\mathbf{\hat{y}}$ at position $j$, and $|\mathcal{V}|$ is the size of the vocabulary in the LLM.

For the entire dataset $\mathcal{D}$, the total loss is calculated as the average of all sample losses:
\begin{equation}
\mathcal{L}_{\rm SFT} = \frac{1}{N}\sum_{i=1}^N \ell_{\rm CE}(\mathbf{y}_i, \mathcal{M}_{\theta}(\mathbf{x}_i))
\end{equation}

\noindent\textbf{Translate-then-SFT.} For the translation-then-SFT paradigm, we define the machine translation system as a function $\mathcal{T}$, which accepts text in one language as the source language (\texttt{Src}) and outputs equivalent text in the target language (\texttt{Tgt}). Using the machine translation system $\mathcal{T}$, each pair $(\mathbf{x}_i, \mathbf{y}_i)$ is translated into the target language, resulting in the translated dataset $\mathcal{D}^{\rm MT} = \{(\mathbf{x}^{\rm MT}_i, \mathbf{y}^{\rm MT}_i)\}_{i=1}^N = \{\mathcal{T}(\mathbf{x}_i, \mathbf{y}_i)\}_{i=1}^N$.

Similar to Eq.~\ref{eq:sft}, the LLM $\mathcal{M}_{\theta}$ is then trained on the translated dataset $\mathcal{D}'$, where the loss for a single sample $(\mathbf{x}^{\rm MT}, \mathbf{y}^{\rm MT})$ is computed as:
\begin{equation}
\ell_{\rm CE}(\mathbf{y}^{\rm MT}, \mathbf{\hat{y}}^{\rm MT}) = -\sum_{j=1}^{|\mathcal{V}|} y^{\rm MT}_j \log(\hat{y}^{\rm MT}_j)
\end{equation}
where $\mathbf{\hat{y}}^{\rm MT} = \mathcal{M}_{\theta}(\mathbf{x^{\rm MT}})$ is the response of models to the question $\mathbf{x^{\rm MT}}$ in target language.

\subsection{Self-Distillation from Resource-Rich Languages (SDRRL)}
LLMs exhibit superior comprehension and generation capabilities in resource-rich languages, which we suppose can be a learning reference for other languages to enhance the multilingual capabilities of LLMs. To achieve this, we propose sentence-level knowledge distillation from resource-rich language responses. The core motivation is that the responses of LLMs in the resource-rich language serve as samples from the resource-rich language representation space. By adding these responses and their translations to the transfer set, the gap for cross-linguistic learning is reduced, facilitating the improvement of multilingual capabilities.

\subsubsection{Transfer Set Construction}
We construct a transfer set for sentence-level distillation \textit{by collecting LLM responses in the resource-rich language}. For the original instruction dataset $\mathcal{D} = \{(\mathbf{x}_i, \mathbf{y}_i)\}_{i=1}^N$, LLM $\mathcal{M}_{\theta}$ generates responses for each question $\mathbf{x}_i$, yielding $\mathbf{\hat{y}}_i = \mathcal{M}_{\theta}(\mathbf{x}_i)$, then we get the generated dataset $\mathcal{G} = \{(\mathbf{x}_i, \mathbf{\hat{y}}_i)\}_{i=1}^N = \{(\mathbf{x}_i, \mathcal{M}_{\theta}(\mathbf{x}_i))\}_{i=1}^N$. The synthesized transfer set $\mathcal{D}_{\text{synth}}$ is obtained by equally probable random sampling from both datasets $\mathcal{D}$ and $\mathcal{G}$:
\begin{equation}
\mathcal{D}_{\text{synth}} = \text{Sample}(\mathcal{D}) \cup \text{Sample}(\mathcal{G})
\end{equation}

\subsubsection{Transfer Set Translation}
The above constructed transfer set $\mathcal{D}_{\text{synth}}$ contains question $\mathbf{x}_i$, ground-truth answer $\mathbf{y}_i$, and response $\mathbf{\hat{y}}_i$ by LLM $\mathcal{M}_{\theta}$. We consider translating them into the target language using the machine translation system $\mathcal{T}$, resulting in $\mathbf{x}^{\rm MT}_i = \mathcal{T}(\mathbf{x}_i), \mathbf{y}^{\rm MT}_i = \mathcal{T}(\mathbf{y}_i)$, and $\mathbf{\hat{y}}^{\rm MT}_i = \mathcal{T}(\mathbf{\hat{y}}_i)$. Moreover, we use WMT22-cometkiwi-da~\cite{rei-etal-2022-cometkiwi} as a reference-free metric to assess the translation quality where the translation quality with scores below a threshold $\tau=0.8$ is rejected.

In particular, four sub-datasets are generated, each containing different language combinations of questions and responses:
\begin{itemize}
    \setlength\itemsep{0em}
    \item $\mathcal{D}_{\rm LL}$: Both the questions and responses remain in the resource-rich language, \textit{i.e.}, $\{\mathbf{x}_i,\mathbf{y}_i\}$ or $\{\mathbf{x}_i,\hat{\mathbf{y}}_i\}$.
    \item $\mathcal{D}_{\rm TL}$: The questions are translated into the target language, while responses remain in the resource-rich language, \textit{i.e.}, $\{\mathcal{T}(\mathbf{x}_i), \mathbf{y}_i\}$ or $\{\mathcal{T}({\mathbf{x}}_i), \hat{\mathbf{y}}_i\}$.
    \item $\mathcal{D}_{\rm LT}$: The questions remain in the resource-rich language, while responses are translated into the target language, \textit{i.e.}, $\{\mathbf{x}_i, \mathcal{T}(\mathbf{y}_i)\}$ or $\{\mathbf{x}_i, \mathcal{T}(\hat{\mathbf{y}}_i)\}$.
    \item $\mathcal{D}_{\rm TT}$: Both the questions and responses are translated into the target language, \textit{i.e.}, $\{\mathcal{T}(\mathbf{x}_i), \mathcal{T}(\mathbf{y}_i)\}$ or $\{\mathcal{T}(\mathbf{x}_i), \mathcal{T}(\hat{\mathbf{y}}_i)\}$.
\end{itemize}

This approach, by providing semantically identical but linguistically diverse samples, aids in the implicit alignment of language representation spaces, enhancing unified multilingual performance. Furthermore, $\mathcal{D}_{\rm TL}$ and $\mathcal{D}_{\rm LT}$ enhance LLM's cross-linguistic generative capabilities, helping mitigate off-target issues in target language generation.

\subsubsection{Applying Code-Switching}
Through the aforementioned machine translation process, we achieve alignment in sentence level (\textit{i.e.}, the sentence of question-answer pairs). Additionally, token-level alignment is introduced using a code-switching tool, applied only to the question components $\mathbf{x}_i$ of $\mathcal{D}_{\rm LL}$, $\mathcal{D}_{\rm TL}$, $\mathcal{D}_{\rm LT}$, and $\mathcal{D}_{\rm TT}$ to increase language diversity without compromising generative capabilities.

Specifically, given $\mathbf{x}_i$ composed of a sequence of tokens $\mathbf{x}_i = {x}_{i,1}, {x}_{i,2}, \ldots, {x}_{i,n}$, where ${x}_{i,k}$ denotes the $k$-th token in question $\mathbf{x}_i$ (similarly for $\mathbf{\hat{x}}^{\rm MT}_i$), the code-switched version ${x}_{i,k}$ for each token is generated by applying the rule:
\begin{equation}
{x}_{i,k} = 
\begin{cases} 
\text{Dict}({x}_{i,k}) & \text{with probability}\ p; \\
{x}_{i,k} & \text{with probability}\ 1-p,
\end{cases}
\end{equation}
where each token ${x}_{i,k}$ in $\mathbf{x}_i$ is replaced by its corresponding token in the bilingual dictionary for code-switching $\text{Dict}({x}_{i,k})$ with a $p=0.15$ probability if ${x}_{i,k}$ is found in the bilingual dictionary.  Responses, either in the source language $\mathbf{y}_i$ (similarly for $\mathbf{\hat{y}}_i$) or the target language $\mathbf{y}_i^{\rm MT}$ (similarly for $\mathbf{\hat{y}}_i^{\rm MT}$), remain unchanged.


\subsubsection{Incorporating External Parallel Corpus}

\begin{table}[h!]
    \centering
    \resizebox{\columnwidth}{!}{
    \begin{tabular}{p{\columnwidth}}
    \toprule
    \textbf{The Template for Constructing $\mathcal{D}_{\text{mt}}$ and $\mathcal{D}_{\text{comp}}$} \\
    \midrule
    \textcolor{gray}{\# Construct Data for Machine Translation} \\
    Question: Translate the following sentence from \textit{English} to \textit{Indonesian}.\\
    \textit{The quick brown fox jumps over the lazy dog.}\\
    Answer: \textit{Sang rubah cokelat cepat melompati anjing malas.}\\
    \\
    \textcolor{gray}{\# Construct Data for Sentence Completion} \\
    Question: Complete the following sentence in \textit{Indonesian} according to its context.\\
    \textit{Sang rubah cokelat cepat}
    \\
    Answer: 
    \textit{Sang rubah cokelat cepat melompati anjing malas.}
    \\
    \bottomrule
    \end{tabular}}
    \caption{The template for constructing $\mathcal{D}_{\text{mt}}$ and $\mathcal{D}_{\text{comp}}$ with Indonesian-English as an example.  $\mathcal{D}_{\text{mt}}$ includes bidirectional translations.  $\mathcal{D}_{\text{comp}}$ contains only the target language sentences, which are split at random positions. }
    \label{table:add_mt_data}
\end{table}

The target language sequences $\mathbf{\hat{y}}_i$ synthesized by the external machine translation system $\mathcal{T}$ may contain low-quality translations, thereby introducing a significant amount of noise into the knowledge distillation transfer dataset $\mathcal{D}_{\text{synth}}$.To mitigate the impact of noise on the multilingual generative capabilities of LLMs, we leverage a tiny external parallel corpus $\mathcal{P} = \{ (\mathbf{s}_i,\mathbf{t}_i)^L_{i=1}\}$ between the resource-rich language \texttt{Src} and the target language \texttt{Tgt}. Based on the templates in Table~\ref{table:add_mt_data}, we can construct two parts of instruction data: machine translation task instructions ($\mathcal{D}_{\text{mt}}$) and sentence completion task instructions ($\mathcal{D}_{\text{comp}}$). By incorporating these two parts, the transfer set includes non-synthetic natural target language texts, which helps improve the generative quality of LLMs in the target language.

\subsubsection{Training Objective}
The final training dataset $\mathcal{D}_{\text{train}}$ includes $\mathcal{D}_{\rm LL}$, $\mathcal{D}_{\rm TL}$, $\mathcal{D}_{\rm LT}$, $\mathcal{D}_{\rm TT}$, $\mathcal{D}_{\text{mt}}$, and $\mathcal{D}_{\text{comp}}$. The total loss function is defined as:
\begin{equation}
\mathcal{L}_{\rm SDRRL}=\sum_{\substack{d \in \mathcal{U}}}\frac{1}{|\mathcal{D}_d|}\sum_{\{\mathbf{x}, \mathbf{y}\} \in \mathcal{D}_d}\ell_{\rm CE}(\mathcal{M}_{\theta}(\mathbf{x}), \mathbf{y}),
\end{equation}
where $\mathcal{U}= \{\rm LL, TL, LT, TT, mt, comp\}$  and $\mathcal{D}_d$ corresponds to the respective data subset (\textit{e.g.}, $\mathcal{D}_{\rm LL}$, $\mathcal{D}_{\rm TL}$, \textit{etc}.).

\section{Experiments}
\subsection{Setup}
We use LLaMA-2-7B~\cite{touvron2023llama2} as the base model. Drawing from the distribution of language in pre-training corpus, we use English (\textsc{Eng}) as a resource-rich language and conduct experiments on 14 languages: Czech (\textsc{Ces}), Danish (\textsc{Dan}), Ukrainian (\textsc{Ukr}), Bulgarian (\textsc{Bul}), Finnish (\textsc{Fin}), Hungarian (\textsc{Hun}), Norwegian (\textsc{Nob}), Indonesian (\textsc{Ind}), Japanese (\textsc{Jpn}), Korean (\textsc{Kor}), Portuguese (\textsc{Por}), Slovenian (\textsc{Slv)}, Vietnamese (\textsc{Vie}), and Polish (\textsc{Pol}). Stanford Alpaca instruction data~\cite{alpaca} serve as the base of the transfer set $\mathcal{D}$, providing questions and ground-truth answers in English. For machine translation, we utilize open-source NLLB-200-3.3B~\cite{costa2022no} model. To improve translation quality, we follow~\citet{zeng-etal-2021-wechat} to filter low-quality translations and use CLD3~\cite{cld3} model to remove off-target translations. We also follow~\citet{lin2021pretraining} to construct bilingual dictionaries for code-switching. See appendix~\ref{app:implement} for more details.

\paragraph{Implementation Details} Our code is implemented using DeepSpeed~\cite{deepspeed} on eight NVIDIA A800-SXM4-80GB GPUs. Following~\citet{7wang2023openchat}, we set the training duration to four epochs with an automatically calculated learning rate and employ early stopping. Other hyperparameters are set according to~\citet{llama-factory}.

\paragraph{Baselines.} For comparison, we consider the following baseline systems that enhance LLaMA-2's multilingual capabilities using different instruction fine-tuning methods:
\begin{itemize}
    \setlength\itemsep{0em}
    \item \textbf{SFT}~\cite{ouyang2022training}: It only involves English instruction datasets in the process of fine-tuning, which is not multilingual-oriented.
    \item \textbf{Translate-then-SFT}~\cite[T-SFT]{35chen2023breaking}: It uses an external machine translation system to translate English instruction data into non-English languages and construct multilingual data for instruction fine-tuning.
    \item \textbf{Cross-Lingual Instruction Tuning} (CIT;~\citealp{23li2023align}): It constructs cross-lingual instructions for fine-tuning, imposing models to respond in the target language given the source language as context.
    \item \textbf{Cross-Lingual Chain-of-Thought Reasoning} (XCOT;~\citealp{22chai2024xcot}): It applies code-switching to multilingual instruction training data, using high-resource instruction data to supervise the training of low-resource languages with cross-lingual distillation.
\end{itemize}

\paragraph{Datasets.} We evaluate the multilingual capabilities of LLMs on four representative datasets:
\begin{itemize}
    \setlength\itemsep{0em}
    \item \textbf{BELEBELE}~\cite{9bandarkar2023belebele}: A widely used language understanding dataset covering 122 languages, where each question, linked to a short passage, has four multiple-choice answers. This dataset proves challenging for state-of-the-art LLMs. Accuracy is reported in our experiments.
    \item \textbf{FLORES}~\cite{goyal-etal-2022-flores}: A benchmark for machine translation with parallel text from Wikipedia for 204 languages, making up over 40,000 directions. We evaluate the bidirectional translation results between the target language and English, reporting scores using SacreBLEU~\cite{post-2018-call} and COMET score using WMT22-comet-da model~\cite{rei-etal-2022-comet}.
    \item \textbf{XL-SUM}~\cite{hasan-etal-2021-xl}: A multilingual abstractive summarization benchmark for 44 languages, comprising multiple long news texts requiring summarization into a single sentence. ROUGE-1 and ROUGE-L F1 scores are reported.
    \item \textbf{MKQA}~\cite{mkqa}: An open-domain question-answering dataset across 26 diverse languages, providing multiple possible short answers as ground truth for each question. We use the official evaluation script and report token overlapped F1 scores. 
\end{itemize}

\begin{table*}[t!]
\centering
\scalebox{0.76}{
  \begin{tabular}{lcccccccccccccccc}
    \toprule
    & \textsc{\textbf{Ces}} & \textsc{\textbf{Dan}} & \textsc{\textbf{Ukr}} & \textsc{\textbf{Bul}} & \textsc{\textbf{Fin}} & \textsc{\textbf{Hun}} & \textsc{\textbf{Nob}} & \textsc{\textbf{Ind}} & \textsc{\textbf{Jpn}} & \textsc{\textbf{Kor}} & \textsc{\textbf{Por}} & \textsc{\textbf{Slv}} & \textsc{\textbf{Vie}} & \textsc{\textbf{Pol}} & \textsc{\textbf{Avg.}} \\
    \midrule
    \multicolumn{16}{c}{\textit{Performance on Target Language}} \\
    SFT & 49.33&48.33&46.67&49.11&39.78&43.22&49.22&46.15&42.01&37.99&55.98&42.79&42.91&44.69&45.58  \\
    T-SFT & 48.22&51.67&47.11&51.22&47.11&\underline{45.67}&51.33&49.72&41.56&43.69&56.20&46.03&\underline{47.60}&48.72&48.28  \\
    CIT & 50.11&53.44&47.22&51.44&\underline{48.00}&\underline{45.67}&\underline{53.33}&\underline{49.94}&\underline{43.24}&\underline{46.26}&\underline{56.65}&\underline{46.70}&45.59&\underline{49.72}&\underline{49.09} \\
    XCOT & \underline{51.56}&\underline{54.22}&\underline{47.83}&\underline{52.78}&47.00&\underline{45.67}&51.33&49.16&43.02&46.15&56.42&46.48&46.82&48.49&49.07  \\
    \textbf{SDRRL} & \textbf{52.11}&\textbf{55.00}&\textbf{48.33}&\textbf{54.00}&\textbf{49.56}&\textbf{46.44}&\textbf{53.89}&\textbf{52.40}&\textbf{45.81}&\textbf{46.82}&\textbf{57.88}&\textbf{47.26}&\textbf{48.38}&\textbf{50.17}&\textbf{50.58}  \\
    \midrule
    \multicolumn{16}{c}{\textit{Performance on English Language}} \\
    SFT & 65.39&65.39&65.39&65.39&65.39&65.39&65.39&65.39&\textbf{65.39}&65.39&65.39&65.39&65.39&65.39&\underline{65.39}   \\
    T-SFT & 63.91&65.25&66.03&65.25&65.70&65.36&65.25&65.70&61.01&60.45&63.80&\underline{65.47}&\underline{65.47}&65.92&64.61  \\
    CIT & 63.46&\underline{65.47}&65.59&64.02&61.23&63.46&64.13&\underline{65.92}&62.01&63.46&64.02&63.24&62.91&62.91&63.70  \\
    XCOT & \underline{65.70}&\underline{65.47}&\underline{66.15}&\underline{66.48}&\underline{65.81}&\underline{65.70}&\underline{66.55}&64.92&63.24&\underline{65.43}&62.46&66.50&63.91&\underline{66.37}&65.34  \\
    \textbf{SDRRL} & \textbf{66.26}&\textbf{65.70}&\textbf{67.15}&\textbf{66.53}&\textbf{65.92}&\textbf{66.70}&\textbf{66.59}&\textbf{67.15}&\underline{65.13}&\textbf{65.45}&\textbf{66.48}&\textbf{66.59}&\textbf{65.57}&\textbf{66.82}&\textbf{66.29}  \\
    \bottomrule
  \end{tabular}
  }
  \caption{Results of baselines and our SDRRL on BELEBELE benchmark. In each column, the best result is \textbf{in bold} and the second best result is \underline{underlined}.}
  \label{tab:belebele}
\end{table*}

\begin{table*}[t!]
\centering
\scalebox{0.76}{
  \begin{tabular}{lccccccccccccccc}
    \toprule
    & \textsc{\textbf{Ces}} & \textsc{\textbf{Dan}} & \textsc{\textbf{Ukr}} & \textsc{\textbf{Bul}} & \textsc{\textbf{Fin}} & \textsc{\textbf{Hun}} & \textsc{\textbf{Nob}} & \textsc{\textbf{Ind}} & \textsc{\textbf{Jpn}} & \textsc{\textbf{Kor}} & \textsc{\textbf{Por}} & \textsc{\textbf{Slv}} & \textsc{\textbf{Vie}} & \textsc{\textbf{Pol}} & \textsc{\textbf{Avg.}}\\
    \midrule
    \multicolumn{16}{c}{\textit{BLEU scores on X-to-English Tasks}} \\
    SFT & \underline{34.66}&\underline{42.57}&\underline{34.17}&\underline{33.91}&\underline{26.76}&\underline{28.15}&\underline{38.34}&20.78&\phantom{0}7.56&\phantom{0}3.15&33.25&11.94&16.01&15.31&24.75
 \\
    T-SFT & 32.63&32.21&31.13&31.05&23.53&24.18&27.44&23.38&\phantom{0}7.82&\phantom{0}7.68&33.03&14.36&19.63&\underline{19.43}&23.39
\\
    CIT & 26.54&29.88&24.25&26.66&21.51&21.24&30.21&\underline{29.02}&\phantom{0}6.00&\phantom{0}7.58&34.46&16.57&\underline{25.84}&19.19&22.78
 \\
    XCOT & 31.52&31.26&29.90&31.05&24.37&23.60&32.50&27.33&\phantom{0}\underline{8.29}&\phantom{0}\underline{9.23}&\underline{35.86}&\underline{17.82}&25.46&19.40&\underline{24.83}
 \\
    \textbf{SDRRL} & \textbf{36.38}&\textbf{45.71}&\textbf{35.33}&\textbf{37.49}&\textbf{30.80}&\textbf{31.62}&\textbf{40.88}&\textbf{30.93}&\textbf{15.42}&\textbf{12.20}&\textbf{39.81}&\textbf{21.15}&\textbf{28.68}&\textbf{22.52} &\textbf{30.64}\\
    \midrule
    \multicolumn{16}{c}{\textit{BLEU scores on English-to-X Tasks}} \\
    SFT & 13.00&21.91&11.18&12.98&\phantom{0}8.39&\phantom{0}9.07&18.53&34.54&17.03&\underline{18.15}&\underline{43.06}&\underline{28.46}&25.06&\underline{27.65}&20.64
 \\
    T-SFT & 22.68&27.78&\underline{23.11}&\underline{27.59}&15.31&16.96&25.60&\underline{31.79}&\underline{19.52}&18.11&39.75&26.17&25.09&26.04&\underline{24.68}
 \\
    CIT & 22.03&28.57&19.92&26.85&14.54&\underline{17.46}&\underline{25.97}&29.46&13.81&15.33&35.24&22.60&22.33&22.84&22.64
 \\
    XCOT & 2\underline{3.11}&\underline{32.20}&21.97&27.33&\underline{15.80}&17.38&25.96&30.33&\phantom{0}9.31&15.13&38.04&26.56&\underline{25.43}&26.03&23.90 \\
    \textbf{SDRRL} & \textbf{27.91}&\textbf{39.00}&\textbf{27.25}&\textbf{33.93}&\textbf{20.88}&\textbf{22.09}&\textbf{29.64}&\textbf{35.32}&\textbf{20.51}&\textbf{20.47}&\textbf{43.36}&\textbf{30.09}&\textbf{29.87}&\textbf{27.86}&\textbf{29.16} \\
    \midrule
    \midrule
    \multicolumn{16}{c}{\textit{COMET scores on X-to-English Tasks}} \\
    SFT & \underline{85.35}&\underline{87.60}&\underline{84.58}&\underline{84.97}&\underline{85.69}&\underline{84.40}&\underline{86.35}&73.54&63.41&45.44&78.91&80.98&63.43&68.46&76.65
 \\
    T-SFT & 84.71&84.26&83.33&83.82&83.78&82.02&83.31&78.94&78.39&72.95&80.38&\underline{81.82}&73.81&\underline{79.06}&80.76
 \\
    CIT & 81.71&82.84&80.06&81.72&82.14&82.29&83.37&\underline{84.62}&76.16&73.88&\underline{83.71}&76.38&\underline{80.60}&78.97&80.60
 \\
    XCOT & 84.40&84.47&83.11&83.90&84.67&81.96&84.68&83.50&\underline{78.83}&\underline{75.66}&83.23&76.48&79.46&78.75&\underline{81.65}
 \\
    \textbf{SDRRL} & \textbf{86.04}&\textbf{88.51}&\textbf{84.82}&\textbf{86.08}&\textbf{86.98}&\textbf{85.70}&\textbf{87.15}&\textbf{89.46}&\textbf{83.33}&\textbf{79.02}&\textbf{85.15}&\textbf{84.02}&\textbf{81.43}&\textbf{83.08}&\textbf{85.06} \\
    \midrule
    \multicolumn{16}{c}{\textit{COMET scores on English-to-X Tasks}} \\
    SFT & 57.19&70.93&55.25&54.99&60.29&53.94&71.97&83.82&82.46&\underline{82.14}&84.57&55.96&80.78&82.14&69.75
 \\
    T-SFT & 78.94&81.34&\underline{79.92}&81.43&78.53&76.01&82.69&\underline{84.90}&\underline{82.76}&80.58&86.42&69.06&81.62&\underline{82.86}&80.50
 \\
    CIT & \underline{79.87}&82.47&78.63&\underline{81.70}&78.39&\underline{76.18}&83.19&84.18&78.15&78.45&85.12&\underline{80.12}&80.77&80.17&\underline{80.53}
 \\
    XCOT & 79.22&\underline{83.29}&79.16&80.86&\underline{78.63}&75.51&\underline{83.30}&84.68&74.27&77.31&\underline{86.01}&78.65&\underline{82.24}&82.72&80.42
 \\
    \textbf{SDRRL} & \textbf{84.29}&\textbf{86.91}&\textbf{83.51}&\textbf{85.40}&\textbf{84.62}&\textbf{81.06}&\textbf{85.55}&\textbf{86.00}&\textbf{83.65}&\textbf{82.66}&\textbf{87.64}&\textbf{82.63}&\textbf{83.93}&\textbf{83.61}&\textbf{84.39} \\
    \bottomrule
  \end{tabular}
  }
  \caption{Results of baselines and our SDRRL on FLORES benchmark.}
  \label{tab:flores}
\end{table*}

\subsection{Main Results}
Table~\ref{tab:belebele} shows the experimental results of the multilingual understanding task. Tables~\ref{tab:flores}, Table~\ref{tab:xlsum1} and Table~\ref{tab:mkqa} show the results on multilingual generation tasks. From the experimental results, we can observe that:

(1) \textbf{SDRRL effectively enhances performance in the target languages.} Specifically, in every comprehension and generation task, our method surpasses the baselines in almost all target languages. As shown in Table~\ref{tab:belebele}, SDRRL improves [erformane in comprehension tasks by approximately +1.5 BLEU score. On the Flores dataset, SDRRL yields up to about +6.0 BLEU score improvement in both directions and about +4.0 COMET score improvement (Table~\ref{tab:flores}). This demonstrates that using proficient responses in resource-rich languages as supervisory signals for knowledge distillation significantly enhances performance in other target languages.

(2) \textbf{SDRRL exhibits stronger robustness in generation tasks.} For example, on the XL-SUM dataset (Table~\ref{tab:xlsum1}), which requires the generation of longer texts, the average performance of CIT and XCOT decreased due to the quality of machine-translated texts and pipeline noise, yet SDRRL still achieved about +0.55 ROUGE-L F1 score improvement. On the FLORES dataset (Table~\ref{tab:flores}), which requires cross-lingual text generation, T-SFT and CIT lead to decrease of -1.36 and -2.08 BLEU scores, respectively, while our method improves by +5.88 BLEU scores. This suggests that adding machine-translated data constructed instructions to the self-distillation process effectively improves multilingual generation and mitigates the negative impact of low-quality translated texts.

\begin{table}[t!]
\centering
\scalebox{0.71}{
  \begin{tabular}{lccccccc}
    \toprule
    & \textsc{\textbf{Ind}} & \textsc{\textbf{Jpn}} & \textsc{\textbf{Kor}} & \textsc{\textbf{Por}} & \textsc{\textbf{Vie}} & \textsc{\textbf{Ukr}} &\textsc{\textbf{Avg.}} \\
    \midrule
    
    \multicolumn{8}{c}{\textit{Performance on Target Language (ROUGE-1)}} \\
    SFT & 20.82&\phantom{0}6.17&0.66&23.38&\phantom{0}9.30&\phantom{0}8.10&11.41 \\
    T-SFT & \underline{25.61} &32.11 &\underline{7.67} &\underline{26.68} &\underline{20.59} &\underline{14.19} &\underline{21.14} \\
    CIT & 24.64&16.11&5.80&26.33&20.55&11.21 &17.44 \\
    XCOT & 22.55&\underline{32.39}&7.26&26.21&19.84&13.38&20.27 \\
    \textbf{SDRRL} & \textbf{26.08}&\textbf{33.15}&\textbf{8.18}&\textbf{27.40}&\textbf{20.98}&\textbf{14.35}&\textbf{21.69} \\
    \midrule
    \multicolumn{8}{c}{\textit{Performance on Target Language (ROUGE-L)}} \\
    SFT & 16.03&\phantom{0}4.13&0.61&15.84&\phantom{0}7.21&\phantom{0}6.72&\phantom{0}8.42 \\
    T-SFT & \underline{20.15} &\textbf{22.83}&\underline{6.93} &\underline{18.41} &\underline{15.18} &\underline{11.73} &\underline{15.87} \\
    CIT & 19.02&11.22&5.14&18.06&14.91&9.00 &12.89 \\
    XCOT & 17.19&22.32&6.52&18.05&14.57&10.78&14.91 \\
    \textbf{SDRRL} & \textbf{20.47}&\underline{22.81}&\textbf{7.35}&\textbf{19.09}&\textbf{15.52}&\textbf{11.84}&\textbf{16.18} \\
    \midrule \midrule
    \multicolumn{8}{c}{\textit{Performance on English Language (ROUGE-1)}} \\
    SFT & 26.35&26.35&26.35&26.35&26.35&26.35&26.35 \\
    T-SFT & 27.49&26.89&\underline{26.68} &27.28 &26.42&\underline{26.75} &\underline{26.92} \\
    CIT & \underline{27.84}&\underline{27.40}&26.57&\underline{27.39}&\underline{27.17} &24.83 &26.87 \\
    XCOT & 26.44&25.45&25.43&26.78&26.00&25.90&26.00 \\
    \textbf{SDRRL} & \textbf{28.18}&\textbf{27.73 }&\textbf{27.44}&\textbf{27.57}&\textbf{27.52}&\textbf{27.23}&\textbf{27.61}\\
    \midrule 
    \multicolumn{8}{c}{\textit{Performance on English Language (ROUGE-L)}} \\
    SFT & 18.68&18.68&18.68&18.68&18.68&18.68&18.68 \\
    T-SFT & 19.64&19.11&\underline{18.94} &19.54 &18.73&\underline{19.01} &\underline{19.16}\\
    CIT & \underline{19.93}&\underline{19.56}&18.81&\underline{19.56}&\underline{19.34} &17.43 &19.11 \\
    XCOT & 18.63&17.83&17.86&18.91&18.29&18.15&18.28 \\
    \textbf{SDRRL} & \textbf{20.25} &\textbf{19.88 }&\textbf{19.56}&\textbf{19.69}&\textbf{19.66}&\textbf{19.44}&\textbf{19.75} \\
    \bottomrule
  \end{tabular}
  }
  \caption{Results of baselines and our SDRRL on XL-SUM benchmark on the target language and English.}
  \label{tab:xlsum1}
\end{table}

\begin{table}[!t]
\scalebox{0.62}{
\renewcommand\tabcolsep{3pt}
\begin{tabular}{lcccccccccc}
  
    \toprule
    & \textsc{\textbf{Nob}} & \textsc{\textbf{Dan}} & \textsc{\textbf{Fin}} & \textsc{\textbf{Hun}} & \textsc{\textbf{Jpn}} & \textsc{\textbf{Kor}} & \textsc{\textbf{Por}} & \textsc{\textbf{Vie}} & \textsc{\textbf{Pol}}  &\textsc{\textbf{Avg.}} \\
    \midrule
    \multicolumn{11}{c}{\textit{Performance on Target Language}} \\
    SFT&37.30&38.28&37.30&35.21&32.80&33.18&39.29&37.50&37.50&36.48 \\
T-SFT&39.73&39.59&\underline{38.95}&\underline{38.60}&\underline{33.96}&33.90&39.93&38.71&38.14&37.95 \\

CIT&\underline{40.18}&\underline{39.94}&37.94&38.40&33.50&\textbf{34.24}&39.86&\underline{39.94}&\underline{38.84}&\underline{38.09} \\

XCOT&39.03&39.28&38.12&35.60&33.07&33.69&\underline{39.96}&39.49&38.49&37.41 \\

\textbf{SDRRL}&\textbf{40.64}&\textbf{40.92}&\textbf{39.71}&\textbf{39.02}&\textbf{39.51}&\underline{34.06}&\textbf{41.12}&\textbf{40.02}&\textbf{39.45}&\textbf{39.38}
 \\
 \midrule \midrule
    \multicolumn{11}{c}{\textit{Performance on English Language}} \\
    SFT&41.62&41.62&41.62&41.62&41.62&41.62&41.62&41.62&41.62&41.62 \\

T-SFT&\underline{44.92}&42.63&\underline{44.24}&\underline{44.21}&41.65&42.11&42.63&\underline{42.65}&42.81&43.09 \\

CIT&44.09&\underline{43.86}&43.55&44.12&\underline{42.83}&\underline{43.29}&42.51&42.52&\underline{43.41}&\underline{43.35} \\

XCOT&43.23&43.16&43.53&43.06&42.59&42.58&\underline{43.39}&42.58&43.29&43.05 \\

\textbf{SDRRL}&\textbf{45.42}&\textbf{45.33}&\textbf{45.47}&\textbf{44.78}&\textbf{43.26}&\textbf{43.58}&\textbf{43.99}&\textbf{45.77}&\textbf{44.71}&\textbf{44.70}
 \\
    \bottomrule
    
  \end{tabular}}
\caption{Results of baselines and our SDRRL on MKQA dataset on the target language and English.}
\label{tab:mkqa}
\end{table}

(3) \textbf{SDRRL can maintain the original strong capabilities in English.} The results show that it is more challenging to retain the original English capabilities for languages with unique alphabets (\textit{e.g.}, Japanese and Korean). For example, in the Japanese comprehension task (Table~\ref{tab:belebele}), all baseline methods lead to a performance drop in English compared to SFT, while only our method successfully preserving the original English capabilities.

\subsection{Ablation Study}
\begin{table}[!t]
  \centering
  \scalebox{0.8}{
    \begin{tabular}{llcccc}
    \toprule
    & & \multicolumn{2}{c}{\textsc{\textbf{NLU Avg.}}} & \multicolumn{2}{c}{\textsc{\textbf{NLG Avg.}}} \\
    \cmidrule(lr){3-4} \cmidrule(lr){5-6}
    & {\bf  } & {\textsc{\textbf{Tar.}}} & {\textsc{\textbf{Eng}}} & {\textsc{\textbf{Tar.}}} & {\textsc{\textbf{Eng}}} \\
    \midrule
    1 & Full Method & {\bf 50.58} & {\bf 66.29} & {\bf 28.24} & {\bf 31.69} \\
    2 & \quad -  $\mathcal{D}_{\rm TL}$ and $\mathcal{D}_{\rm LT}$ &49.56&65.93&26.15&30.55 \\
    3 & \quad - $\mathcal{D}_{\rm synth}$ + $\mathcal{D}$&48.59&65.10&25.16&30.10 \\
    4 & \quad -  $\mathcal{D}_{\rm mt}$ and $\mathcal{D}_{\rm comp}$ & \underline{50.41} &\underline{66.01}&26.61&30.19 \\
    5 & \quad -  Code Switching & 50.37&65.94&\underline{27.13}&\underline{30.69} \\
    6 & Only $\mathcal{D}_{\rm mt}$ and $\mathcal{D}_{\rm comp}$  & 41.25&61.61&17.89&22.28 \\
    \bottomrule
    \end{tabular}
    }
      \caption{Ablation study. Average scores of target language (\textsc{Tar}.) and English (\textsc{Eng}) on natural language understanding task (NLU, including BELEBELE) and natural language generation tasks (NLG, including FLORES, XL-SUM ROUGE-L, and MKQA) are reported.}
      \label{tab:sub2}
\end{table}

We further investigate the
effectiveness of each component of our proposed SDRRL.  The results are shown in Table~\ref{tab:sub2}, where average scores on  natural language understanding and generation tasks are reported. Our observations include:

(1) Rows 1 to 5 demonstrate that removing any single component results in performance degradation, affirming the necessity and efficacy of each component in SDRRL.

(2) Insights from row 3 suggest a significant performance decline in both target languages and English when model-generated responses ($\hat{\textbf{y}}_i$) are removed from $\mathcal{D}_{\rm synth}$, highlighting the effectiveness of utilizing responses in resource-rich languages as additional supervision signals for improving multilingual capabilities. Moreover, row 2 indicates that substituting sentences with their semantic counterparts in different languages also contributes to multilingual performance improvement.

(3) Row 4 and 5 reveal that $\mathcal{D}_{\rm mt}$, $\mathcal{D}_{\rm comp}$, and code-switching provide a limited amount of ground truth. This additional supervisory signal is beneficial for generative tasks and helps improve the quality of responses.

(4) Despite introducing a small amount of parallel data through $\mathcal{D}_{\rm mt}$ and $\mathcal{D}_{\rm comp}$, as shown in row 6, relying solely on these additional data for LLM training supervision leads to severe performance degradation. Compared to row 4, this indicates that these data do not inherently bring positive performance gains but are used to mitigate the deterioration of the LLM's multilingual generative representation space caused by noisy machine-translated text, serving as a regularization mechanism in knowledge distillation.

\subsection{Visualization of Representation Space for Source and Target Langauges}
\begin{figure}[t!]
    \includegraphics[width=1.0\columnwidth]{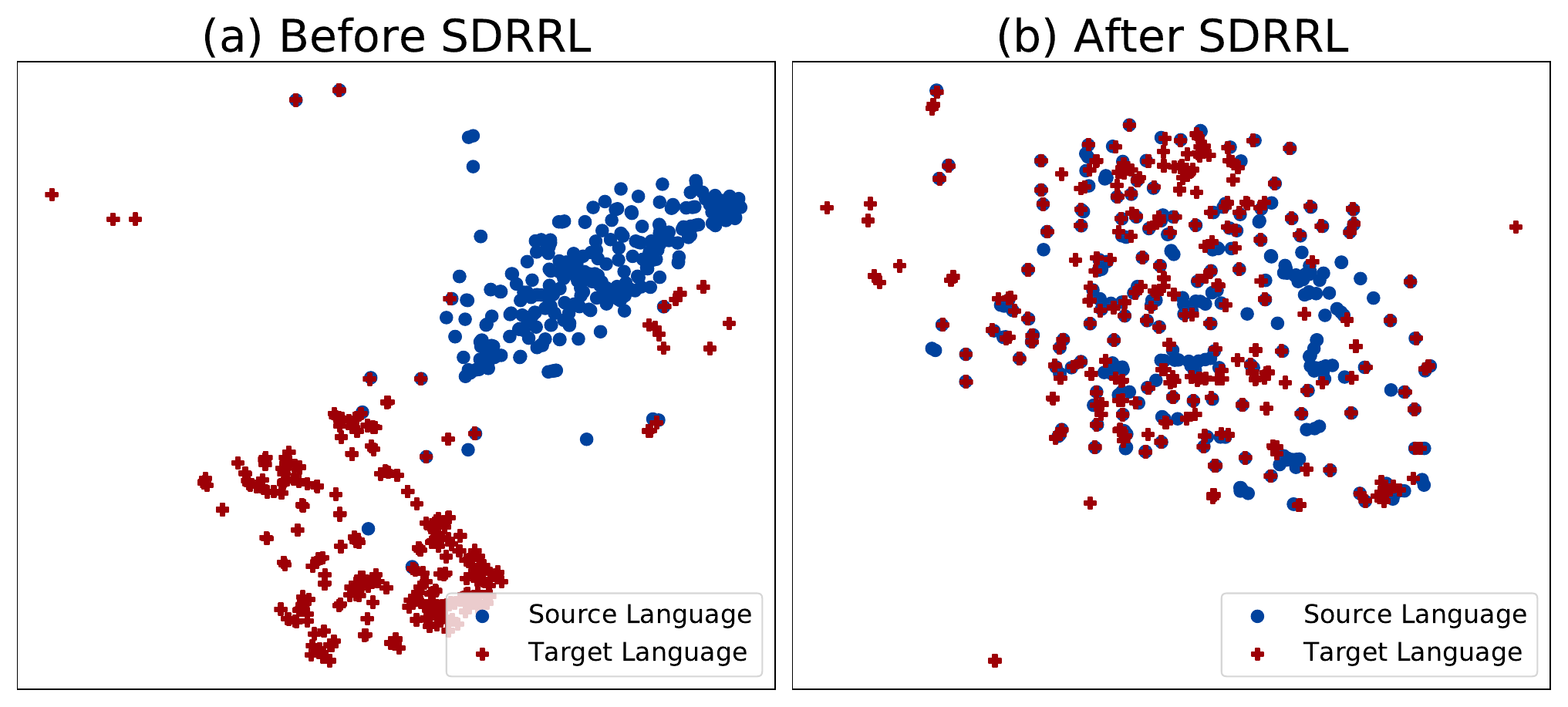}
    \caption{t-SNE visualizations of output representations by LLaMA-2 before and after applying SDRRL. The markers in red and blue represent semantically equivalent instructions in different languages.}
    \label{fig:tsne}
\end{figure}
We visualize the sentence representations of input instructions to investigate the effect of SDRRL on the multilingual representation space. Following common practices in sequence classification~\citet{li2023label}, we input instructions into the LLaMA-2 and use the last hidden states of the last token as the vector representation of the sentence. We then apply t-SNE~\cite{van2008visualizing} to reduce the 4096-dim representations to 2-dim for visualization. 

As shown in Figure~\ref{fig:tsne}, after applying SDRRL, the representations of semantically equivalent instructions in the source and target languages are drawn closer together. This implies that SDRRL has improved the multilingual representation space by aligning the representation space of the target language closer to that of the resource-rich, better-modeled source language, thereby enhancing the performance in target languages.

\subsection{SeaLLM as Different Backbone Model}
By using the responses of LLMs in high-resource languages as the supervisory signal for knowledge distillation, SDRRL is applicable to various LLMs, not limited to LLaMA-2. In this part, we conduct experiments on SeaLLM-7B~\cite{nguyen2023seallms}, a specialized language model optimized for Southeast Asian languages. 

As shown in Table~\ref{tab:main_sea}, SDRRL results in an improvement of +2.39 average scores on the target languages. In English, SDRRL maintains its original performance, while the baselines exhibit a performance drop of at least -2.02 average scores compared to vanilla SFT. This demonstrates the generalizability of SDRRL in different LLMs. See appendix~\ref{app:south} for detailed results on more datasets.

\subsection{Further Analysis}
\paragraph{Non-English Source Languages.} SDRRL is also capable of transferring multilingual performance using other source languages in high-resource. In appendix~\ref{app:french}, we opt for experiments with French instead of English. Experimental results reveal that, despite the LLM and the machine translation system exhibiting stronger performance in English, SDRRL still achieves positive distillation gains with French as the source language.

\paragraph{Case Study.} 
In appendix~\ref{app:case}, we provide several case studies to offer deeper insights into the impact of SDRRL on the generation capabilities of LLMs. It is observed that the SDRRL process is able to alleviate off-target issues in the target language, reduce grammatical errors and hallucinations, and enhance the fluency of the output text.

\begin{table}[!t]
  \centering
  \scalebox{0.76}{
  \begin{threeparttable}
    \renewcommand\tabcolsep{5.0pt}
    \begin{tabular}{lccccc}
    \toprule
 & {\textsc{\textbf{BELE.}}} & {\textsc{\textbf{XL-SUM}}}  & {\textsc{\textbf{FLORES}}}& {\textsc{\textbf{MKQA}}}& {\textsc{\textbf{Avg.}}}\\
    \midrule
    \multicolumn{6}{c}{\textit{Performance on Target Language}}\\
    SFT &42.24&\underline{16.48}&18.45&38.86&29.01 \\
     T-SFT &\underline{42.77}&15.32&16.59&43.40&29.52 \\
      CIT &42.53&15.75&\underline{20.49}&\underline{43.70}&\underline{30.62} \\
       XCOT &41.19&15.79&17.21&42.04&29.06 \\
        \textbf{SDRRL} &\textbf{43.67}&\textbf{17.89}&\textbf{25.86}&\textbf{44.63}&\textbf{33.01} \\
        \midrule \midrule
        \multicolumn{6}{c}{\textit{Performance on English Language}}\\
         SFT &\underline{60.19}&15.25&\underline{28.49}&\underline{39.62}&\underline{35.89} \\
     T-SFT &58.70&\underline{15.63}&23.72&37.43&33.87 \\
      CIT &58.66&15.42&18.31&36.67&32.27 \\
       XCOT &57.73&14.90&23.96&37.94&33.63 \\
        \textbf{SDRRL} &\textbf{60.67}&\textbf{16.24}&\textbf{29.47}&\textbf{40.32}&\textbf{36.68} \\
    \bottomrule
    \end{tabular}
    \end{threeparttable}}
      \caption{Results of baselines and our SDRRL on SeaLLM. The average scores across various datasets are reported, and full results are available in appendix~\ref{app:south}.}
      \label{tab:main_sea}
\end{table}
\section{Conclusion and Future Work}
We introduce Self-Distillation from Resource-Rich Languages (SDRRL) to enhance the multilingual capabilities of LLMs. SDRRL uses the model itself to generate high-quality responses in resource-rich source languages and their target language counterparts as supervision signals for knowledge distillation, aiming to align additional target languages with resource-rich languages. We conduct comprehensive experiments across 16 languages on LLaMA-2 and SeaLLM. The results demonstrate that, compared to various baselines, our method significantly enhances the performance of target languages while preserving the capabilities of source languages. This highlights the multilingual potential of LLMs and illuminates paths for further research into multilingual LLMs.


\section*{Limitations}
Firstly, within our method pipeline, some components are interchangeable. For example, our approach relies on external machine translation systems to provide translations in the target language, while future research could explore self-translation with LLMs that achieve great low-resource translation capabilities, thereby simplifying the process. Additionally, our method uses a small amount of machine-translated parallel corpus to construct the transfer set, but employing monolingual texts in the target language represents a promising research direction. Secondly, our experiments are conducted with only a single source language and target language. Subsequent research could investigate using a mix of multiple languages as both source and target languages and explore the mutual influences between different languages to further enhance the multilingual capabilities of LLMs. Thirdly, our method does not involve engineering on the architecture of LLMs. For specific extremely low-resource languages, modifying the architecture and introducing additional data, such as vocabulary expansion or continuing pre-training, might be beneficial in enhancing multilingual performance.
\nocite{*}
\bibliography{anthology,custom}

\begin{thebibliography}{75}
\expandafter\ifx\csname natexlab\endcsname\relax\def\natexlab#1{#1}\fi

\bibitem[{Bandarkar et~al.(2023)Bandarkar, Liang, Muller, Artetxe, Shukla, Husa, Goyal, Krishnan, Zettlemoyer, and Khabsa}]{9bandarkar2023belebele}
Lucas Bandarkar, Davis Liang, Benjamin Muller, Mikel Artetxe, Satya~Narayan Shukla, Donald Husa, Naman Goyal, Abhinandan Krishnan, Luke Zettlemoyer, and Madian Khabsa. 2023.
\newblock \href {http://arxiv.org/abs/2308.16884} {The belebele benchmark: a parallel reading comprehension dataset in 122 language variants}.

\bibitem[{Chai et~al.(2024)Chai, Yang, Sun, Guo, Liu, Wang, Liang, Bai, Li, Peng, and Li}]{22chai2024xcot}
Linzheng Chai, Jian Yang, Tao Sun, Hongcheng Guo, Jiaheng Liu, Bing Wang, Xiannian Liang, Jiaqi Bai, Tongliang Li, Qiyao Peng, and Zhoujun Li. 2024.
\newblock \href {http://arxiv.org/abs/2401.07037} {xcot: Cross-lingual instruction tuning for cross-lingual chain-of-thought reasoning}.

\bibitem[{Chen et~al.(2023)Chen, Zheng, Wu, Gong, Song, Zhang, and Li}]{35chen2023breaking}
Nuo Chen, Zinan Zheng, Ning Wu, Ming Gong, Yangqiu Song, Dongmei Zhang, and Jia Li. 2023.
\newblock \href {http://arxiv.org/abs/2310.20246} {Breaking language barriers in multilingual mathematical reasoning: Insights and observations}.

\bibitem[{Chen et~al.(2020)Chen, Gan, Cheng, Liu, and Liu}]{40chen2020distilling}
Yen-Chun Chen, Zhe Gan, Yu~Cheng, Jingzhou Liu, and Jingjing Liu. 2020.
\newblock \href {http://arxiv.org/abs/1911.03829} {Distilling knowledge learned in bert for text generation}.

\bibitem[{Chi et~al.(2021)Chi, Dong, Zheng, Huang, Mao, Huang, and Wei}]{27chi2021improving}
Zewen Chi, Li~Dong, Bo~Zheng, Shaohan Huang, Xian-Ling Mao, Heyan Huang, and Furu Wei. 2021.
\newblock \href {http://arxiv.org/abs/2106.06381} {Improving pretrained cross-lingual language models via self-labeled word alignment}.

\bibitem[{Conover et~al.(2023)Conover, Hayes, Mathur, Xie, Wan, Shah, Ghodsi, Wendell, Zaharia, and Xin}]{DatabricksBlog2023DollyV2}
Mike Conover, Matt Hayes, Ankit Mathur, Jianwei Xie, Jun Wan, Sam Shah, Ali Ghodsi, Patrick Wendell, Matei Zaharia, and Reynold Xin. 2023.
\newblock \href {https://www.databricks.com/blog/2023/04/12/dolly-first-open-commercially-viable-instruction-tuned-llm} {Free dolly: Introducing the world's first truly open instruction-tuned llm}.

\bibitem[{Costa-juss{\`a} et~al.(2022)Costa-juss{\`a}, Cross, {\c{C}}elebi, Elbayad, Heafield, Heffernan, Kalbassi, Lam, Licht, Maillard et~al.}]{costa2022no}
Marta~R Costa-juss{\`a}, James Cross, Onur {\c{C}}elebi, Maha Elbayad, Kenneth Heafield, Kevin Heffernan, Elahe Kalbassi, Janice Lam, Daniel Licht, Jean Maillard, et~al. 2022.
\newblock No language left behind: Scaling human-centered machine translation.
\newblock \emph{arXiv preprint arXiv:2207.04672}.

\bibitem[{Etxaniz et~al.(2023)Etxaniz, Azkune, Soroa, de~Lacalle, and Artetxe}]{17etxaniz2023multilingual}
Julen Etxaniz, Gorka Azkune, Aitor Soroa, Oier~Lopez de~Lacalle, and Mikel Artetxe. 2023.
\newblock \href {http://arxiv.org/abs/2308.01223} {Do multilingual language models think better in english?}

\bibitem[{Google et~al.(2023)Google, Anil, Borgeaud, Wu, Alayrac, Yu, Soricut, Schalkwyk, Dai, Hauth et~al.}]{team2023gemini}
Gemini~Team Google, Rohan Anil, Sebastian Borgeaud, Yonghui Wu, Jean-Baptiste Alayrac, Jiahui Yu, Radu Soricut, Johan Schalkwyk, Andrew~M Dai, Anja Hauth, et~al. 2023.
\newblock Gemini: a family of highly capable multimodal models.
\newblock \emph{arXiv preprint arXiv:2312.11805}.

\bibitem[{Gordon and Duh(2019)}]{39gordon2019explaining}
Mitchell~A. Gordon and Kevin Duh. 2019.
\newblock \href {http://arxiv.org/abs/1912.03334} {Explaining sequence-level knowledge distillation as data-augmentation for neural machine translation}.

\bibitem[{Gou et~al.(2021)Gou, Yu, Maybank, and Tao}]{45gou2021knowledge}
Jianping Gou, Baosheng Yu, Stephen~J Maybank, and Dacheng Tao. 2021.
\newblock Knowledge distillation: A survey.
\newblock \emph{International Journal of Computer Vision}, 129:1789--1819.

\bibitem[{Goyal et~al.(2022)Goyal, Gao, Chaudhary, Chen, Wenzek, Ju, Krishnan, Ranzato, Guzm{\'a}n, and Fan}]{goyal-etal-2022-flores}
Naman Goyal, Cynthia Gao, Vishrav Chaudhary, Peng-Jen Chen, Guillaume Wenzek, Da~Ju, Sanjana Krishnan, Marc{'}Aurelio Ranzato, Francisco Guzm{\'a}n, and Angela Fan. 2022.
\newblock \href {https://doi.org/10.1162/tacl_a_00474} {The {F}lores-101 evaluation benchmark for low-resource and multilingual machine translation}.
\newblock \emph{Transactions of the Association for Computational Linguistics}, 10:522--538.

\bibitem[{Hasan et~al.(2021)Hasan, Bhattacharjee, Islam, Mubasshir, Li, Kang, Rahman, and Shahriyar}]{hasan-etal-2021-xl}
Tahmid Hasan, Abhik Bhattacharjee, Md.~Saiful Islam, Kazi Mubasshir, Yuan-Fang Li, Yong-Bin Kang, M.~Sohel Rahman, and Rifat Shahriyar. 2021.
\newblock \href {https://doi.org/10.18653/v1/2021.findings-acl.413} {{XL}-sum: Large-scale multilingual abstractive summarization for 44 languages}.
\newblock In \emph{Findings of the Association for Computational Linguistics: ACL-IJCNLP 2021}, pages 4693--4703, Online. Association for Computational Linguistics.

\bibitem[{Hinton et~al.(2015)Hinton, Vinyals, and Dean}]{37hinton2015distilling}
Geoffrey Hinton, Oriol Vinyals, and Jeff Dean. 2015.
\newblock \href {http://arxiv.org/abs/1503.02531} {Distilling the knowledge in a neural network}.

\bibitem[{Hiyouga(2023)}]{llama-factory}
Hiyouga. 2023.
\newblock Llama factory.
\newblock \url{https://github.com/hiyouga/LLaMA-Factory}.

\bibitem[{Huang et~al.(2023)Huang, Tang, Zhang, Zhao, Song, Xia, and Wei}]{18huang2023languages}
Haoyang Huang, Tianyi Tang, Dongdong Zhang, Wayne~Xin Zhao, Ting Song, Yan Xia, and Furu Wei. 2023.
\newblock \href {http://arxiv.org/abs/2305.07004} {Not all languages are created equal in llms: Improving multilingual capability by cross-lingual-thought prompting}.

\bibitem[{Huang et~al.(2022)Huang, Ma, Zhang, Wei, and Wang}]{13huang2022zeroshot}
Lianzhe Huang, Shuming Ma, Dongdong Zhang, Furu Wei, and Houfeng Wang. 2022.
\newblock \href {http://arxiv.org/abs/2202.11451} {Zero-shot cross-lingual transfer of prompt-based tuning with a unified multilingual prompt}.

\bibitem[{Jiang et~al.(2023)Jiang, Sablayrolles, Mensch, Bamford, Chaplot, Casas, Bressand, Lengyel, Lample, Saulnier et~al.}]{jiang2023mistral}
Albert~Q Jiang, Alexandre Sablayrolles, Arthur Mensch, Chris Bamford, Devendra~Singh Chaplot, Diego de~las Casas, Florian Bressand, Gianna Lengyel, Guillaume Lample, Lucile Saulnier, et~al. 2023.
\newblock Mistral 7b.
\newblock \emph{arXiv preprint arXiv:2310.06825}.

\bibitem[{Jiang et~al.(2024)Jiang, Sablayrolles, Roux, Mensch, Savary, Bamford, Chaplot, de~las Casas, Hanna, Bressand, Lengyel, Bour, Lample, Lavaud, Saulnier, Lachaux, Stock, Subramanian, Yang, Antoniak, Scao, Gervet, Lavril, Wang, Lacroix, and Sayed}]{8jiang2024mixtral}
Albert~Q. Jiang, Alexandre Sablayrolles, Antoine Roux, Arthur Mensch, Blanche Savary, Chris Bamford, Devendra~Singh Chaplot, Diego de~las Casas, Emma~Bou Hanna, Florian Bressand, Gianna Lengyel, Guillaume Bour, Guillaume Lample, Lélio~Renard Lavaud, Lucile Saulnier, Marie-Anne Lachaux, Pierre Stock, Sandeep Subramanian, Sophia Yang, Szymon Antoniak, Teven~Le Scao, Théophile Gervet, Thibaut Lavril, Thomas Wang, Timothée Lacroix, and William~El Sayed. 2024.
\newblock \href {http://arxiv.org/abs/2401.04088} {Mixtral of experts}.

\bibitem[{Kew et~al.(2023)Kew, Schottmann, and Sennrich}]{34kew2023turning}
Tannon Kew, Florian Schottmann, and Rico Sennrich. 2023.
\newblock \href {http://arxiv.org/abs/2312.12683} {Turning english-centric llms into polyglots: How much multilinguality is needed?}

\bibitem[{Kim and Rush(2016)}]{38kim2016sequencelevel}
Yoon Kim and Alexander~M. Rush. 2016.
\newblock \href {http://arxiv.org/abs/1606.07947} {Sequence-level knowledge distillation}.

\bibitem[{Lample and Conneau(2019)}]{2lample2019crosslingual}
Guillaume Lample and Alexis Conneau. 2019.
\newblock \href {http://arxiv.org/abs/1901.07291} {Cross-lingual language model pretraining}.

\bibitem[{Li et~al.(2023{\natexlab{a}})Li, Wang, Zhang, and Zong}]{23li2023align}
Chong Li, Shaonan Wang, Jiajun Zhang, and Chengqing Zong. 2023{\natexlab{a}}.
\newblock \href {http://arxiv.org/abs/2311.08089} {Align after pre-train: Improving multilingual generative models with cross-lingual alignment}.

\bibitem[{Li et~al.(2023{\natexlab{b}})Li, Koto, Wu, Aji, and Baldwin}]{12li2023bactrianx}
Haonan Li, Fajri Koto, Minghao Wu, Alham~Fikri Aji, and Timothy Baldwin. 2023{\natexlab{b}}.
\newblock \href {http://arxiv.org/abs/2305.15011} {Bactrian-x: Multilingual replicable instruction-following models with low-rank adaptation}.

\bibitem[{Li et~al.(2022)Li, Tang, Zhao, Nie, and Wen}]{1li2022pretrained}
Junyi Li, Tianyi Tang, Wayne~Xin Zhao, Jian-Yun Nie, and Ji-Rong Wen. 2022.
\newblock \href {http://arxiv.org/abs/2201.05273} {Pretrained language models for text generation: A survey}.

\bibitem[{Li et~al.(2023{\natexlab{c}})Li, Li, Liu, Xie, Li, lee Wang, Li, and Zhong}]{li2023label}
Zongxi Li, Xianming Li, Yuzhang Liu, Haoran Xie, Jing Li, Fu~lee Wang, Qing Li, and Xiaoqin Zhong. 2023{\natexlab{c}}.
\newblock \href {http://arxiv.org/abs/2310.01208} {Label supervised llama finetuning}.

\bibitem[{Lin et~al.(2022)Lin, Mihaylov, Artetxe, Wang, Chen, Simig, Ott, Goyal, Bhosale, Du, Pasunuru, Shleifer, Koura, Chaudhary, O'Horo, Wang, Zettlemoyer, Kozareva, Diab, Stoyanov, and Li}]{4lin2022fewshot}
Xi~Victoria Lin, Todor Mihaylov, Mikel Artetxe, Tianlu Wang, Shuohui Chen, Daniel Simig, Myle Ott, Naman Goyal, Shruti Bhosale, Jingfei Du, Ramakanth Pasunuru, Sam Shleifer, Punit~Singh Koura, Vishrav Chaudhary, Brian O'Horo, Jeff Wang, Luke Zettlemoyer, Zornitsa Kozareva, Mona Diab, Veselin Stoyanov, and Xian Li. 2022.
\newblock \href {http://arxiv.org/abs/2112.10668} {Few-shot learning with multilingual language models}.

\bibitem[{Lin et~al.(2021)Lin, Pan, Wang, Qiu, Feng, Zhou, and Li}]{lin2021pretraining}
Zehui Lin, Xiao Pan, Mingxuan Wang, Xipeng Qiu, Jiangtao Feng, Hao Zhou, and Lei Li. 2021.
\newblock \href {http://arxiv.org/abs/2010.03142} {Pre-training multilingual neural machine translation by leveraging alignment information}.

\bibitem[{Longpre et~al.(2020)Longpre, Lu, and Daiber}]{mkqa}
Shayne Longpre, Yi~Lu, and Joachim Daiber. 2020.
\newblock \href {https://arxiv.org/pdf/2007.15207.pdf} {Mkqa: A linguistically diverse benchmark for multilingual open domain question answering}.

\bibitem[{Ma et~al.(2022)Ma, Nguyen, and Ma}]{14}
Yukun Ma, Trung~Hieu Nguyen, and Bin Ma. 2022.
\newblock \href {https://doi.org/10.1109/ICASSP43922.2022.9746935} {Cpt: Cross-modal prefix-tuning for speech-to-text translation}.
\newblock In \emph{ICASSP 2022 - 2022 IEEE International Conference on Acoustics, Speech and Signal Processing (ICASSP)}, pages 6217--6221.

\bibitem[{Mao and Yu(2024{\natexlab{a}})}]{30mao2024tuning}
Zhuoyuan Mao and Yen Yu. 2024{\natexlab{a}}.
\newblock \href {http://arxiv.org/abs/2401.05811} {Tuning llms with contrastive alignment instructions for machine translation in unseen, low-resource languages}.

\bibitem[{Mao and Yu(2024{\natexlab{b}})}]{20mao2024tuning}
Zhuoyuan Mao and Yen Yu. 2024{\natexlab{b}}.
\newblock \href {http://arxiv.org/abs/2401.05811} {Tuning llms with contrastive alignment instructions for machine translation in unseen, low-resource languages}.

\bibitem[{Nguyen et~al.(2023)Nguyen, Zhang, Li, Aljunied, Tan, Cheng, Chen, Deng, Yang, Liu, Zhang, and Bing}]{nguyen2023seallms}
Xuan-Phi Nguyen, Wenxuan Zhang, Xin Li, Mahani Aljunied, Qingyu Tan, Liying Cheng, Guanzheng Chen, Yue Deng, Sen Yang, Chaoqun Liu, Hang Zhang, and Lidong Bing. 2023.
\newblock \href {http://arxiv.org/abs/2312.00738} {Seallms -- large language models for southeast asia}.

\bibitem[{Niklaus et~al.(2023)Niklaus, Matoshi, Rani, Galassi, Stürmer, and Chalkidis}]{10Niklaus_2023}
Joel Niklaus, Veton Matoshi, Pooja Rani, Andrea Galassi, Matthias Stürmer, and Ilias Chalkidis. 2023.
\newblock \href {https://doi.org/10.18653/v1/2023.findings-emnlp.200} {Lextreme: A multi-lingual and multi-task benchmark for the legal domain}.
\newblock In \emph{Findings of the Association for Computational Linguistics: EMNLP 2023}. Association for Computational Linguistics.

\bibitem[{Ooms(2024)}]{cld3}
Jeroen Ooms. 2024.
\newblock \href {https://docs.ropensci.org/cld3/ https://github.com/ropensci/cld3 https://ropensci.r-universe.dev/cld3} {\emph{cld3: Google's Compact Language Detector 3}}.
\newblock R package version 1.6.0.

\bibitem[{OpenAI(2022)}]{openaichatgpt}
OpenAI. 2022.
\newblock Chat{GPT}.
\newblock \url{https://openai.com/chatgpt}.

\bibitem[{OpenAI(2023)}]{openaigpt4}
OpenAI. 2023.
\newblock \href {http://arxiv.org/abs/2303.08774} {{GPT}-4 technical report}.

\bibitem[{Ouyang et~al.(2022)Ouyang, Wu, Jiang, Almeida, Wainwright, Mishkin, Zhang, Agarwal, Slama, Ray et~al.}]{ouyang2022training}
Long Ouyang, Jeffrey Wu, Xu~Jiang, Diogo Almeida, Carroll Wainwright, Pamela Mishkin, Chong Zhang, Sandhini Agarwal, Katarina Slama, Alex Ray, et~al. 2022.
\newblock Training language models to follow instructions with human feedback.
\newblock \emph{Advances in Neural Information Processing Systems}, 35:27730--27744.

\bibitem[{Pahune and Chandrasekharan(2023)}]{16Pahune_2023}
Saurabh Pahune and Manoj Chandrasekharan. 2023.
\newblock \href {https://doi.org/10.22214/ijraset.2023.54677} {Several categories of large language models (llms): A short survey}.
\newblock \emph{International Journal for Research in Applied Science and Engineering Technology}, 11(7):615–633.

\bibitem[{Pham et~al.(2022)Pham, Cho, Joshi, and Hegde}]{43pham2022revisiting}
Minh Pham, Minsu Cho, Ameya Joshi, and Chinmay Hegde. 2022.
\newblock \href {http://arxiv.org/abs/2206.08491} {Revisiting self-distillation}.

\bibitem[{Post(2018)}]{post-2018-call}
Matt Post. 2018.
\newblock \href {https://doi.org/10.18653/v1/W18-6319} {A call for clarity in reporting {BLEU} scores}.
\newblock In \emph{Proceedings of the Third Conference on Machine Translation: Research Papers}, pages 186--191, Brussels, Belgium. Association for Computational Linguistics.

\bibitem[{Qin et~al.(2023)Qin, Chen, Wei, Huang, and Che}]{33qin2023crosslingual}
Libo Qin, Qiguang Chen, Fuxuan Wei, Shijue Huang, and Wanxiang Che. 2023.
\newblock \href {http://arxiv.org/abs/2310.14799} {Cross-lingual prompting: Improving zero-shot chain-of-thought reasoning across languages}.

\bibitem[{Ranaldi and Pucci(2023)}]{32ranaldi2023does}
Leonardo Ranaldi and Giulia Pucci. 2023.
\newblock Does the english matter? elicit cross-lingual abilities of large language models.
\newblock In \emph{Proceedings of the 3rd Workshop on Multi-lingual Representation Learning (MRL)}, pages 173--183.

\bibitem[{Ranaldi and Zanzotto(2023)}]{19ranaldi2023empowering}
Leonardo Ranaldi and Fabio~Massimo Zanzotto. 2023.
\newblock \href {http://arxiv.org/abs/2311.08097} {Empowering multi-step reasoning across languages via tree-of-thoughts}.

\bibitem[{Rasley et~al.(2020)Rasley, Rajbhandari, Ruwase, and He}]{deepspeed}
Jeff Rasley, Samyam Rajbhandari, Olatunji Ruwase, and Yuxiong He. 2020.
\newblock \href {https://doi.org/10.1145/3394486.3406703} {Deepspeed: System optimizations enable training deep learning models with over 100 billion parameters}.
\newblock In \emph{Proceedings of the 26th ACM SIGKDD International Conference on Knowledge Discovery \& Data Mining}, KDD '20, page 3505–3506, New York, NY, USA. Association for Computing Machinery.

\bibitem[{Rei et~al.(2022{\natexlab{a}})Rei, C.~de Souza, Alves, Zerva, Farinha, Glushkova, Lavie, Coheur, and Martins}]{rei-etal-2022-comet}
Ricardo Rei, Jos{\'e}~G. C.~de Souza, Duarte Alves, Chrysoula Zerva, Ana~C Farinha, Taisiya Glushkova, Alon Lavie, Luisa Coheur, and Andr{\'e} F.~T. Martins. 2022{\natexlab{a}}.
\newblock \href {https://aclanthology.org/2022.wmt-1.52} {{COMET}-22: Unbabel-{IST} 2022 submission for the metrics shared task}.
\newblock In \emph{Proceedings of the Seventh Conference on Machine Translation (WMT)}, pages 578--585, Abu Dhabi, United Arab Emirates (Hybrid). Association for Computational Linguistics.

\bibitem[{Rei et~al.(2022{\natexlab{b}})Rei, Treviso, Guerreiro, Zerva, Farinha, Maroti, C.~de Souza, Glushkova, Alves, Coheur, Lavie, and Martins}]{rei-etal-2022-cometkiwi}
Ricardo Rei, Marcos Treviso, Nuno~M. Guerreiro, Chrysoula Zerva, Ana~C Farinha, Christine Maroti, Jos{\'e}~G. C.~de Souza, Taisiya Glushkova, Duarte Alves, Luisa Coheur, Alon Lavie, and Andr{\'e} F.~T. Martins. 2022{\natexlab{b}}.
\newblock \href {https://aclanthology.org/2022.wmt-1.60} {{C}omet{K}iwi: {IST}-unbabel 2022 submission for the quality estimation shared task}.
\newblock In \emph{Proceedings of the Seventh Conference on Machine Translation (WMT)}, pages 634--645, Abu Dhabi, United Arab Emirates (Hybrid). Association for Computational Linguistics.

\bibitem[{Saberi et~al.(2024)Saberi, Fard, and Chen}]{11saberi2024advfusion}
Iman Saberi, Fatemeh Fard, and Fuxiang Chen. 2024.
\newblock \href {http://arxiv.org/abs/2307.07854} {Advfusion: Multilingual adapter-based knowledge transfer for code summarization}.

\bibitem[{Schuster et~al.(2019{\natexlab{a}})Schuster, Ram, Barzilay, and Globerson}]{26schuster2019crosslingual}
Tal Schuster, Ori Ram, Regina Barzilay, and Amir Globerson. 2019{\natexlab{a}}.
\newblock \href {http://arxiv.org/abs/1902.09492} {Cross-lingual alignment of contextual word embeddings, with applications to zero-shot dependency parsing}.

\bibitem[{Schuster et~al.(2019{\natexlab{b}})Schuster, Ram, Barzilay, and Globerson}]{36schuster2019crosslingual}
Tal Schuster, Ori Ram, Regina Barzilay, and Amir Globerson. 2019{\natexlab{b}}.
\newblock \href {http://arxiv.org/abs/1902.09492} {Cross-lingual alignment of contextual word embeddings, with applications to zero-shot dependency parsing}.

\bibitem[{Shaham et~al.(2024)Shaham, Herzig, Aharoni, Szpektor, Tsarfaty, and Eyal}]{25shaham2024multilingual}
Uri Shaham, Jonathan Herzig, Roee Aharoni, Idan Szpektor, Reut Tsarfaty, and Matan Eyal. 2024.
\newblock \href {http://arxiv.org/abs/2401.01854} {Multilingual instruction tuning with just a pinch of multilinguality}.

\bibitem[{Sun et~al.(2020)Sun, Wang, Chen, Utiyama, Sumita, and Zhao}]{44sun2020knowledge}
Haipeng Sun, Rui Wang, Kehai Chen, Masao Utiyama, Eiichiro Sumita, and Tiejun Zhao. 2020.
\newblock \href {http://arxiv.org/abs/2004.10171} {Knowledge distillation for multilingual unsupervised neural machine translation}.

\bibitem[{Taori et~al.(2023)Taori, Gulrajani, Zhang, Dubois, Li, Guestrin, Liang, and Hashimoto}]{alpaca}
Rohan Taori, Ishaan Gulrajani, Tianyi Zhang, Yann Dubois, Xuechen Li, Carlos Guestrin, Percy Liang, and Tatsunori~B. Hashimoto. 2023.
\newblock Stanford alpaca: An instruction-following llama model.
\newblock \url{https://github.com/tatsu-lab/stanford_alpaca}.

\bibitem[{Touvron et~al.(2023{\natexlab{a}})Touvron, Lavril, Izacard, Martinet, Lachaux, Lacroix, Rozière, Goyal, Hambro, Azhar, Rodriguez, Joulin, Grave, and Lample}]{touvron2023llama}
Hugo Touvron, Thibaut Lavril, Gautier Izacard, Xavier Martinet, Marie-Anne Lachaux, Timothée Lacroix, Baptiste Rozière, Naman Goyal, Eric Hambro, Faisal Azhar, Aurelien Rodriguez, Armand Joulin, Edouard Grave, and Guillaume Lample. 2023{\natexlab{a}}.
\newblock \href {http://arxiv.org/abs/2302.13971} {Llama: Open and efficient foundation language models}.

\bibitem[{Touvron et~al.(2023{\natexlab{b}})Touvron, Martin, Stone, Albert, Almahairi, Babaei, Bashlykov, Batra, Bhargava, Bhosale, Bikel, Blecher, Ferrer, Chen, Cucurull, Esiobu, Fernandes, Fu, Fu, Fuller, Gao, Goswami, Goyal, Hartshorn, Hosseini, Hou, Inan, Kardas, Kerkez, Khabsa, Kloumann, Korenev, Koura, Lachaux, Lavril, Lee, Liskovich, Lu, Mao, Martinet, Mihaylov, Mishra, Molybog, Nie, Poulton, Reizenstein, Rungta, Saladi, Schelten, Silva, Smith, Subramanian, Tan, Tang, Taylor, Williams, Kuan, Xu, Yan, Zarov, Zhang, Fan, Kambadur, Narang, Rodriguez, Stojnic, Edunov, and Scialom}]{touvron2023llama2}
Hugo Touvron, Louis Martin, Kevin Stone, Peter Albert, Amjad Almahairi, Yasmine Babaei, Nikolay Bashlykov, Soumya Batra, Prajjwal Bhargava, Shruti Bhosale, Dan Bikel, Lukas Blecher, Cristian~Canton Ferrer, Moya Chen, Guillem Cucurull, David Esiobu, Jude Fernandes, Jeremy Fu, Wenyin Fu, Brian Fuller, Cynthia Gao, Vedanuj Goswami, Naman Goyal, Anthony Hartshorn, Saghar Hosseini, Rui Hou, Hakan Inan, Marcin Kardas, Viktor Kerkez, Madian Khabsa, Isabel Kloumann, Artem Korenev, Punit~Singh Koura, Marie-Anne Lachaux, Thibaut Lavril, Jenya Lee, Diana Liskovich, Yinghai Lu, Yuning Mao, Xavier Martinet, Todor Mihaylov, Pushkar Mishra, Igor Molybog, Yixin Nie, Andrew Poulton, Jeremy Reizenstein, Rashi Rungta, Kalyan Saladi, Alan Schelten, Ruan Silva, Eric~Michael Smith, Ranjan Subramanian, Xiaoqing~Ellen Tan, Binh Tang, Ross Taylor, Adina Williams, Jian~Xiang Kuan, Puxin Xu, Zheng Yan, Iliyan Zarov, Yuchen Zhang, Angela Fan, Melanie Kambadur, Sharan Narang, Aurelien Rodriguez, Robert Stojnic, Sergey Edunov, and Thomas
  Scialom. 2023{\natexlab{b}}.
\newblock \href {http://arxiv.org/abs/2307.09288} {Llama 2: Open foundation and fine-tuned chat models}.

\bibitem[{Van~der Maaten and Hinton(2008)}]{van2008visualizing}
Laurens Van~der Maaten and Geoffrey Hinton. 2008.
\newblock Visualizing data using t-sne.
\newblock \emph{Journal of machine learning research}, 9(11).

\bibitem[{Wang et~al.(2023{\natexlab{a}})Wang, Cheng, Zhan, Li, Song, and Liu}]{7wang2023openchat}
Guan Wang, Sijie Cheng, Xianyuan Zhan, Xiangang Li, Sen Song, and Yang Liu. 2023{\natexlab{a}}.
\newblock \href {http://arxiv.org/abs/2309.11235} {Openchat: Advancing open-source language models with mixed-quality data}.

\bibitem[{Wang et~al.(2023{\natexlab{b}})Wang, Cheng, Zhan, Li, Song, and Liu}]{wang2023openchat}
Guan Wang, Sijie Cheng, Xianyuan Zhan, Xiangang Li, Sen Song, and Yang Liu. 2023{\natexlab{b}}.
\newblock Openchat: Advancing open-source language models with mixed-quality data.
\newblock \emph{arXiv preprint arXiv:2309.11235}.

\bibitem[{Wei et~al.(2022)Wei, Wang, Schuurmans, Bosma, Xia, Chi, Le, Zhou et~al.}]{wei2022chain}
Jason Wei, Xuezhi Wang, Dale Schuurmans, Maarten Bosma, Fei Xia, Ed~Chi, Quoc~V Le, Denny Zhou, et~al. 2022.
\newblock Chain-of-thought prompting elicits reasoning in large language models.
\newblock \emph{Advances in Neural Information Processing Systems}, 35:24824--24837.

\bibitem[{Wen-Yi and Mimno(2023)}]{31Wen_Yi_2023}
Andrea Wen-Yi and David Mimno. 2023.
\newblock \href {https://doi.org/10.18653/v1/2023.emnlp-main.71} {Hyperpolyglot llms: Cross-lingual interpretability in token embeddings}.
\newblock In \emph{Proceedings of the 2023 Conference on Empirical Methods in Natural Language Processing}. Association for Computational Linguistics.

\bibitem[{Wolf et~al.(2020)Wolf, Debut, Sanh, Chaumond, Delangue, Moi, Cistac, Rault, Louf, Funtowicz, Davison, Shleifer, von Platen, Ma, Jernite, Plu, Xu, Le~Scao, Gugger, Drame, Lhoest, and Rush}]{wolf-etal-2020-transformers}
Thomas Wolf, Lysandre Debut, Victor Sanh, Julien Chaumond, Clement Delangue, Anthony Moi, Pierric Cistac, Tim Rault, Remi Louf, Morgan Funtowicz, Joe Davison, Sam Shleifer, Patrick von Platen, Clara Ma, Yacine Jernite, Julien Plu, Canwen Xu, Teven Le~Scao, Sylvain Gugger, Mariama Drame, Quentin Lhoest, and Alexander Rush. 2020.
\newblock \href {https://doi.org/10.18653/v1/2020.emnlp-demos.6} {Transformers: State-of-the-art natural language processing}.
\newblock In \emph{Proceedings of the 2020 Conference on Empirical Methods in Natural Language Processing: System Demonstrations}, pages 38--45, Online. Association for Computational Linguistics.

\bibitem[{Workshop et~al.(2022)Workshop, Scao, Fan, Akiki, Pavlick, Ili{\'c}, Hesslow, Castagn{\'e}, Luccioni, Yvon et~al.}]{3workshop2023bloom}
BigScience Workshop, Teven~Le Scao, Angela Fan, Christopher Akiki, Ellie Pavlick, Suzana Ili{\'c}, Daniel Hesslow, Roman Castagn{\'e}, Alexandra~Sasha Luccioni, Fran{\c{c}}ois Yvon, et~al. 2022.
\newblock Bloom: A 176b-parameter open-access multilingual language model.

\bibitem[{Xue et~al.(2021)Xue, Constant, Roberts, Kale, Al-Rfou, Siddhant, Barua, and Raffel}]{5xue2021mt5}
Linting Xue, Noah Constant, Adam Roberts, Mihir Kale, Rami Al-Rfou, Aditya Siddhant, Aditya Barua, and Colin Raffel. 2021.
\newblock \href {http://arxiv.org/abs/2010.11934} {mt5: A massively multilingual pre-trained text-to-text transformer}.

\bibitem[{Yang et~al.(2019)Yang, Xie, Su, and Yuille}]{46Yang_2019_CVPR}
Chenglin Yang, Lingxi Xie, Chi Su, and Alan~L. Yuille. 2019.
\newblock Snapshot distillation: Teacher-student optimization in one generation.
\newblock In \emph{Proceedings of the IEEE/CVF Conference on Computer Vision and Pattern Recognition (CVPR)}.

\bibitem[{Yang et~al.(2023)Yang, Li, Zhang, and Zong}]{15yang2023bigtranslate}
Wen Yang, Chong Li, Jiajun Zhang, and Chengqing Zong. 2023.
\newblock \href {http://arxiv.org/abs/2305.18098} {Bigtranslate: Augmenting large language models with multilingual translation capability over 100 languages}.

\bibitem[{Yoon et~al.(2024)Yoon, Jang, Kim, Kim, Shafayat, and Seo}]{24yoon2024langbridge}
Dongkeun Yoon, Joel Jang, Sungdong Kim, Seungone Kim, Sheikh Shafayat, and Minjoon Seo. 2024.
\newblock \href {http://arxiv.org/abs/2401.10695} {Langbridge: Multilingual reasoning without multilingual supervision}.

\bibitem[{Zeng et~al.(2021)Zeng, Liu, Li, Ran, Meng, Li, Xu, and Zhou}]{zeng-etal-2021-wechat}
Xianfeng Zeng, Yijin Liu, Ernan Li, Qiu Ran, Fandong Meng, Peng Li, Jinan Xu, and Jie Zhou. 2021.
\newblock \href {https://aclanthology.org/2021.wmt-1.23} {{W}e{C}hat neural machine translation systems for {WMT}21}.
\newblock In \emph{Proceedings of the Sixth Conference on Machine Translation}, pages 243--254, Online. Association for Computational Linguistics.

\bibitem[{Zhang et~al.(2020)Zhang, Williams, Titov, and Sennrich}]{zhang2020improving}
Biao Zhang, Philip Williams, Ivan Titov, and Rico Sennrich. 2020.
\newblock \href {http://arxiv.org/abs/2004.11867} {Improving massively multilingual neural machine translation and zero-shot translation}.

\bibitem[{Zhang et~al.(2022{\natexlab{a}})Zhang, Bao, and Ma}]{47}
L.~Zhang, C.~Bao, and K.~Ma. 2022{\natexlab{a}}.
\newblock \href {https://doi.org/10.1109/TPAMI.2021.3067100} {Self-distillation: Towards efficient and compact neural networks}.
\newblock \emph{IEEE Transactions on Pattern Analysis; Machine Intelligence}, 44(08):4388--4403.

\bibitem[{Zhang et~al.(2022{\natexlab{b}})Zhang, Bao, and Ma}]{42}
Linfeng Zhang, Chenglong Bao, and Kaisheng Ma. 2022{\natexlab{b}}.
\newblock \href {https://doi.org/10.1109/TPAMI.2021.3067100} {Self-distillation: Towards efficient and compact neural networks}.
\newblock \emph{IEEE Transactions on Pattern Analysis and Machine Intelligence}, 44(8):4388--4403.

\bibitem[{Zhang et~al.(2019)Zhang, Song, Gao, Chen, Bao, and Ma}]{41zhang2019teacher}
Linfeng Zhang, Jiebo Song, Anni Gao, Jingwei Chen, Chenglong Bao, and Kaisheng Ma. 2019.
\newblock \href {http://arxiv.org/abs/1905.08094} {Be your own teacher: Improve the performance of convolutional neural networks via self distillation}.

\bibitem[{Zhang et~al.(2023)Zhang, Li, Sun, and Liu}]{zhang2023continual}
Yuanchi Zhang, Peng Li, Maosong Sun, and Yang Liu. 2023.
\newblock \href {http://arxiv.org/abs/2212.09097} {Continual knowledge distillation for neural machine translation}.

\bibitem[{Zhang and Liu(2021)}]{zhang2021directquote}
Yuanchi Zhang and Yang Liu. 2021.
\newblock Directquote: A dataset for direct quotation extraction and attribution in news articles.
\newblock \emph{arXiv preprint arXiv:2110.07827}.

\bibitem[{Zhao et~al.(2024)Zhao, Zhang, Gao, Zhang, Gui, and Huang}]{21zhao2024llama}
Jun Zhao, Zhihao Zhang, Luhui Gao, Qi~Zhang, Tao Gui, and Xuanjing Huang. 2024.
\newblock \href {http://arxiv.org/abs/2401.01055} {Llama beyond english: An empirical study on language capability transfer}.

\bibitem[{Zhu et~al.(2024)Zhu, Huang, Yuan, She, Chen, and Birch}]{28zhu2024question}
Wenhao Zhu, Shujian Huang, Fei Yuan, Shuaijie She, Jiajun Chen, and Alexandra Birch. 2024.
\newblock \href {http://arxiv.org/abs/2401.07817} {Question translation training for better multilingual reasoning}.

\end{thebibliography}
\bibliographystyle{acl_natbib}

\appendix

\section{Implementation Details}
\label{app:implement}

The signature of SacreBLEU we use in this work is ``nrefs:1 | case:mixed | eff:no | tok:flores200 | smooth:exp | version:2.0.0''. The Stanford Alpaca dataset comprises 52,002 entries, licensed under the CC BY-NC 4.0 agreement. For each target language, machine translation parallel corpora are sampled from Opus100~\cite{zhang2020improving}, consisting of 1,000 entries. When utilizing NLLB for machine translation, we set the beam size to 4, with the remaining configurations adopting the default parameters from Huggingface Transformers~\cite{wolf-etal-2020-transformers}. In the reimplementation of baselines, the same machine translation system is employed to provide multilingual alignment data. For the 16 languages involved in our experiments, XL-SUM and MKQA datasets have not covered all of them. During the evaluation of MKQA, questions lacking ground truth are skipped. 

\section{More Detailed Results on SeaLLM}
\label{app:south} We conduct experiments on three common Southeast Asian languages: Indonesian (\textsc{Ind}), Thai (\textsc{Tha}), and Khmer (\textsc{Khm}). As shown in Table~\ref{tab:seallm_belebele},~\ref{tab:seaxlsum1},~\ref{tab:seaflores}, and~\ref{tab:seamkqa}, SDRRL still outperforms the baselines, demonstrating the generalizability of SDRRL in different LLMs.

\begin{table}[htbp!]
\centering\small
\scalebox{0.95}{
  \begin{tabular}{lcccc}
    \toprule
    & \textsc{\textbf{Ind}} & \textsc{\textbf{Khm}} & \textsc{\textbf{Tha}} & \textsc{\textbf{Avg.}} \\
    \midrule
    \multicolumn{5}{c}{\textit{Performance on Target Language}} \\
    SFT&47.71&32.56&46.44&42.24 \\ 
T-SFT&48.31&\underline{32.89}&\underline{46.77}&\underline{42.77} \\ 
CIT&\underline{48.83}&32.56&46.20&42.53 \\ 
XCOT&45.81&32.22&45.55&41.19 \\ 
\textbf{SDRRL}&\textbf{50.39}&\textbf{33.67}&\textbf{46.96}&\textbf{43.67} 
\\
    \midrule
    \multicolumn{5}{c}{\textit{Performance on English Language}} \\
    SFT&60.56&\underline{60.56}&\textbf{59.46}&\underline{60.19} \\ 
T-SFT&\underline{60.78}&57.89&57.43&58.70 \\ 
CIT&58.55&58.10&59.33&58.66 \\ 
XCOT&57.77&57.99&57.43&57.73 \\ 
\textbf{SDRRL}&\textbf{61.68}&\textbf{60.89}&\underline{59.44}&\textbf{60.67}
 \\
    \bottomrule
  \end{tabular}
  }
  \caption{Results of baselines and our SDRRL on BELEBELE benchmark using SeaLLM.}
  \label{tab:seallm_belebele}
\end{table}

\begin{table}[t!]
\centering\small
\scalebox{0.95}{
  \begin{tabular}{lp{1.5cm}p{1.5cm}p{1.5cm}}
    \toprule
    & \textsc{\textbf{Ind}} & \textsc{\textbf{Tha}} &\textsc{\textbf{Avg.}} \\
    \midrule
    
    \multicolumn{4}{c}{\textit{Performance on Target Language (ROUGE-1)}} \\
    SFT&21.91&\underline{24.65}&\underline{23.28} \\ 
T-SFT&21.07&21.26&21.17 \\ 
CIT&21.19&23.93&22.56 \\ 
XCOT&\underline{22.20}&21.85&22.02 \\ 
\textbf{SDRRL}&\textbf{23.62}&\textbf{25.78}&\textbf{24.70}
 \\
    \midrule
    \multicolumn{4}{c}{\textit{Performance on Target Language (ROUGE-L)}} \\
    SFT&16.46&\underline{16.50}&\underline{16.48} \\ 
T-SFT&16.23&14.40&15.32 \\ 
CIT&15.84&15.66&15.75 \\ 
XCOT&\underline{16.93}&14.65&15.79 \\ 
\textbf{SDRRL}&\textbf{18.06}&\textbf{17.73}&\textbf{17.89} 
 \\
\midrule \midrule

 \multicolumn{4}{c}{\textit{Performance on English Language (ROUGE-1)}} \\
    SFT&21.39&22.85&22.12 \\ 
T-SFT&21.93&23.17&\underline{22.55} \\ 
CIT&21.65&23.07&22.36 \\ 
XCOT&21.27&21.99&21.63 \\ 
\textbf{SDRRL}&\textbf{23.47}&\textbf{23.55}&\textbf{23.01}
\\
    \midrule
    \multicolumn{4}{c}{\textit{Performance on English Language (ROUGE-L)}} \\
    SFT&14.79&15.71&15.25 \\ 
T-SFT&\underline{15.19}&16.07&\underline{15.63} \\ 
CIT&14.91&15.92&15.42 \\ 
XCOT&14.66&15.14&14.90 \\ 
\textbf{SDRRL}&\textbf{16.34}&\textbf{16.15}&\textbf{16.24}
 \\
    \bottomrule
  \end{tabular}
  }
  \caption{Results of baselines and our SDRRL on XL-SUM benchmark on the target language using SeaLLM.}
  \label{tab:seaxlsum1}
\end{table}

\begin{table}[t!]
\centering\small
\scalebox{0.95}{
  \begin{tabular}{lcccc}
    \toprule
    & \textsc{\textbf{Ind}} & \textsc{\textbf{Tha}} & \textsc{\textbf{Thm}} & \textsc{\textbf{Avg.}}\\
    \midrule
    \multicolumn{5}{c}{\textit{xx$\rightarrow$en (BLEU)}} \\
   SFT&\underline{36.75}&\underline{20.93}&\underline{20.22}&\underline{28.49} \\ 
T-SFT&32.23&14.41&15.21&23.72 \\ 
CIT&22.52&15.84&14.10&18.31 \\ 
XCOT&33.20&16.48&14.71&23.96 \\ 
\textbf{SDRRL}&\textbf{38.30}&\textbf{21.76}&\textbf{20.64}&\textbf{29.47}
\\
    \midrule
    \multicolumn{5}{c}{\textit{en$\rightarrow$xx (BLEU)}} \\
    SFT&30.26&16.53&6.64&18.45 \\ 
T-SFT&28.29&13.10&4.88&16.59 \\ 
CIT&\underline{31.21}&\underline{18.15}&\underline{9.76}&\underline{20.49} \\ 
XCOT&29.15&14.28&5.26&17.21 \\ 
\textbf{SDRRL}&\textbf{36.28}&\textbf{24.02}&\textbf{15.43}&\textbf{25.86} 
 \\
    \midrule
    \midrule
    \multicolumn{5}{c}{\textit{xx$\rightarrow$en (COMET)}} \\
    SFT&\underline{86.94}&\underline{82.89}&\underline{80.07}&\underline{83.51} \\ 
T-SFT &84.49&74.61&71.29&77.89 \\ 
CIT&80.78&78.87&76.18&78.48 \\ 
COT&85.69&77.57&72.14&78.91 \\ 
\textbf{SDRRL}&\textbf{87.39}&\textbf{83.07}&\textbf{80.63}&\textbf{84.01}
 \\
    \midrule
    \multicolumn{5}{c}{\textit{en$\rightarrow$xx (COMET)}} \\
    SFT&\underline{86.78}&73.22&64.05&75.41 \\ 
T-SFT &85.44&66.95&59.09&72.26 \\ 
CIT&85.80&\underline{74.42}&\underline{69.60}&\underline{77.70} \\ 
COT&85.23&68.26&62.24&73.74 \\ 
\textbf{SDRRL}&\textbf{88.70}&\textbf{79.03}&\textbf{75.97}&\textbf{82.34} 
 \\
    \bottomrule
  \end{tabular}
  }
  \caption{Results of baselines and our SDRRL on FLORES benchmark using SeaLLM.}
  \label{tab:seaflores}
\end{table}

\begin{table}[htbp!]
\centering\small
  \begin{tabular}{lccc}
    \toprule
    & \textsc{\textbf{Tha}} & \textsc{\textbf{Khm}} &\textsc{\textbf{Avg.}} \\
    \midrule
    \multicolumn{4}{c}{\textit{Performance on Target Language}} \\
   SFT&40.68&37.04&38.86 \\ 
T-SFT&48.32&38.48&43.40 \\ 
CIT&\underline{48.38}&\underline{39.01}&\underline{43.70} \\ 
XCOT&45.07&39.00&42.04 \\ 
\textbf{SDRRL}&\textbf{49.44}&\textbf{39.81}&\textbf{44.63}

 \\
 \midrule
    \multicolumn{4}{c}{\textit{Performance on English Language}} \\
    SFT&\underline{39.62}&\underline{39.62}&\underline{39.62} \\ 
T-SFT&37.92&36.94&37.43 \\ 
CIT&37.64&35.69&36.67 \\ 
XCOT&38.40&37.48&37.94 \\ 
\textbf{SDRRL}&\textbf{40.66}&\textbf{39.97}&\textbf{40.32}

 \\
    \bottomrule
  \end{tabular}
  \caption{Results of baselines and our SDRRL on MKQA dataset using SeaLLM.}
  \label{tab:seamkqa}
\end{table}

\section{Experiments with Non-English Language as the Source Language}
\label{app:french}
\begin{table}[!t]
  \centering
   \small
  \begin{threeparttable}
    \renewcommand\tabcolsep{5.0pt}
    \begin{tabular}{lcc}
    \toprule
    & \textsc{\textbf{NLU Avg.}} & {\textsc{\textbf{NLG Avg.}}} \\
    \midrule
     English & {\bf +6.29} &  {\bf +5.31} \\
     French & {+2.94} & {+1.77} \\
    \bottomrule
    \end{tabular}
    \end{threeparttable}
      \caption{Distillation gains from SDRRL with English or French as the source language. Average scores on the natural language understanding task (NLU, including BELEBELE) and natural language generation tasks (NLG, including FLORES, XL-SUM ROUGE-L, and MKQA) are reported.}
      \label{tab:french}
\end{table}

SDRRL aims to transfer the proficiency of LLMs from resource-rich languages to another target language, indicating that the source language is not limited to English. In this section, we use French instead of  English to generate responses and replicate experiments conducted in Indonesian, Japanese, and Korean languages. Table~\ref{tab:french} presents the average knowledge distillation gains from SDRRL over SFT when employing English and French as the source languages, respectively. Due to the inferior performance of LLaMA-2-7B in French~\cite{9bandarkar2023belebele} compared to English and the decreased effectiveness of the external translation system NLLB when French is used as the source language~\cite{costa2022no}, the knowledge distillation gains generated with French as the source language are smaller.

\section{Case Study}
\label{app:case}

\begin{table*}[t]
\centering\small

\renewcommand{\arraystretch}{1.4}
\begin{tabular}{@{} m{1.5cm} p{13.5cm} @{}} 
\toprule

Hungarian & \textbf{Question}: Mit jelent a DNA? \\ & \textbf{Before SDRRL:} {\color{red}A DNA (deoxyribonucleic acid) is the hereditary material present in all living organisms. It is a long double-stranded helix made up of four nitrogenous bases; adenine (A), guanine (G), cytosine (C) and thymine (T). }
\\ & \textbf{After SDRRL:} A DNA a dezoxiribonukleinsav rövidítése, egy molekula, amely két szálból áll, amelyek egymáshoz kötődnek, és hidrogénkötések tartják össze őket. Ez a genetikai anyag, amely az élőlények fejlődését és működését szabályozza, és a sejtek fehérje- és szénhidrátszintézisét irányító utasításokat tartalmazza.
\\
\midrule

Slovenian & \textbf{Question}: Opišite življenje in vladanje kralja Karla II. \\
& \textbf{Before SDRRL:} {\color{red}Kralj Karel II. (1638-1700) je bil   francoski krščanski kralj, ki je vladal od leta 1685 do svoje smrti}. Ko je postal kralj, je zavzel podeželje in se lotil reform, katerek cilj je bil izboljšati upravo in omogočiti zaščito družbe. Karel II. je bil tudi pomorski kralj, {\color{red}ki si je meril svet}\includegraphics[height=1em]{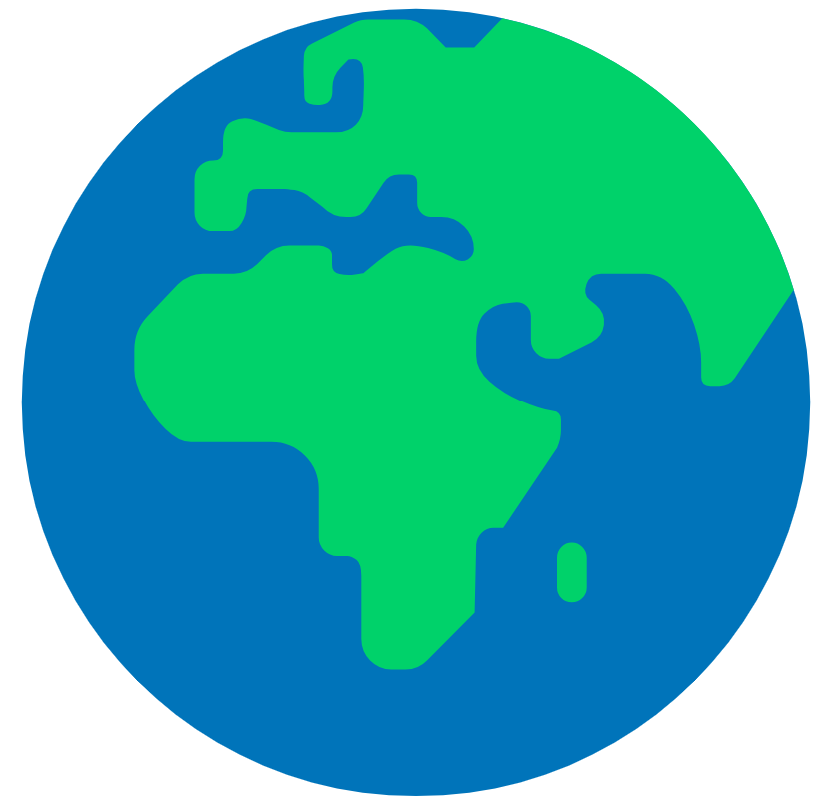}. {\color{red}Under his rule, the French navy became one of the most powerful in the world.} Karel II. je bil tudi velik pokrovitelj umetnosti in književnosti. Že stoletje po njegovem vrnitemenu je postal Karel II. znan kot eno največjih francoskih vladarjev zadnjega obdobja.\\
& \textbf{After SDRRL:} Kralj Karel II. je bil angleški kralj od leta 1660 do svoje smrti leta 1685. Bil je sin Karla I., ki je bil odstranjen z oblasti in usmrčen leta 1649. Po angleški državljanski vojni in obdobju republike je bil Karel II. ponovno kronan za kralja Anglije leta 1660, kar je označilo začetek Kraljeve obnove. Bil je znan kot podpornik umetnosti in kulture, ki je spodbujal delo številnih umetnikov in pisateljev. Njegovo obdobje vladanja je bilo zaznamovano z obnovitvenimi projekti, vključno z razvojem Londona po Velikem požaru leta 1666. Karel II. je umrl leta 1685 brez legitimnih moških potomcev, kar je dolgoročno vplivalo na vprašanje nasledstva. Njegova smrt ni neposredno vplivala na konec Kraljeve obnove ali angleško državljansko vojno, ki se je končala pred njegovim vladanjem, niti na začetek angleške suverenosti v Severni Ameriki.
\\
\midrule
Bulgarian &\textbf{Question}: \selectlanguage{russian}Моля, обяснете историята на ледниковия период.\selectlanguage{english} \\
& \textbf{Before SDRRL:} \selectlanguage{russian}ледников период е период в ерата, в която Земята е {\color{red}bila} предимно покрита с лед. това се дължи на увеличение на {\color{red}азъйствата} на Северния полюс, които се {\color{red}оттикват} водата в океаните и причинява замразяването на течности и земя. това води до увеличение на масата лед {\color{red}вglobus} и {\color{red}снижаваниеa} нива на температурата на {\color{red}аеросферата}. това връвля {\color{red}вanother glacial period or ice age.}\selectlanguage{english} \\
& \textbf{After SDRRL:} \selectlanguage{russian}Ледниковият период е период от време в климатичната история на Земята, през който климатът става много студен и полярните ледени шапки се разширяват, покривайки голяма част от сушата. Това се случва, когато нивото на атмосферната температура значително се понижи, което позволява на водата в океаните да се превърне в лед и води до увеличаване на обема на леда и връщането на вода към морския лед. Ледниковите периоди обикновено са свързани с забележително намаляване на глобалните температури, което се случва в продължение на десетилетия или стотици години. Няколко фактора могат да доведат до ледников период, включително промени в орбитата на Земята, намаляване на вулканичните изригвания и промени в океанските течения. Най-скорошният ледников период, известен като последната ледникова епоха, приключи преди около 10 000 години, когато климатът започна да се затопля и ледниците по света започнаха да се топят.\selectlanguage{english} \\
\bottomrule
\end{tabular}
\caption{Several cases before and after the application of SDRRL. Following SDRRL, the responses from LLMs become smoother, more comprehensive, and contain fewer errors (colored in red).} 
\label{tab:case_st}
\end{table*}
As shown in Table~\ref{tab:case_st}, we provide several cases of SDRRL in several languages. In the case of Hungarian, LLM encounters severe off-target issues, where the response is in English and is inconsistent with the input language Hungarian, which is a frequent problem during multilingual generation. After SDRRL, the severe off-target issue has been effectively mitigated. In the Slovenian case, LLM produces hallucinations when answering factual questions about history, leading to factual inaccuracies, noise tokens like emojis, and off-target English phrases. For example, The time frame mentioned (1685-1700) and the description does not match any king named Charles. It seems there's a mix-up with historical figures. After SDRRL, the hallucination issue has been alleviated, and the generated content becomes more detailed, refined, and fluid. In the Bulgarian scenario, the response contains several grammatical errors, such as ``\selectlanguage{russian}ледников\selectlanguage{english}'', ``\selectlanguage{russian}оттикват\selectlanguage{english}'' and ``\selectlanguage{russian}снижаваниеa\selectlanguage{english}''. In this case, the SDRRL process enhances the clarity and natural flow of the output text while also eliminating grammatical errors in Bulgarian. See appendix~\ref{app:off} for statistical results regarding off-target issues.

\section{Off-Target Issue Analysis}
\label{app:off}
\begin{figure}[t!]
    \includegraphics[width=1.0\columnwidth]{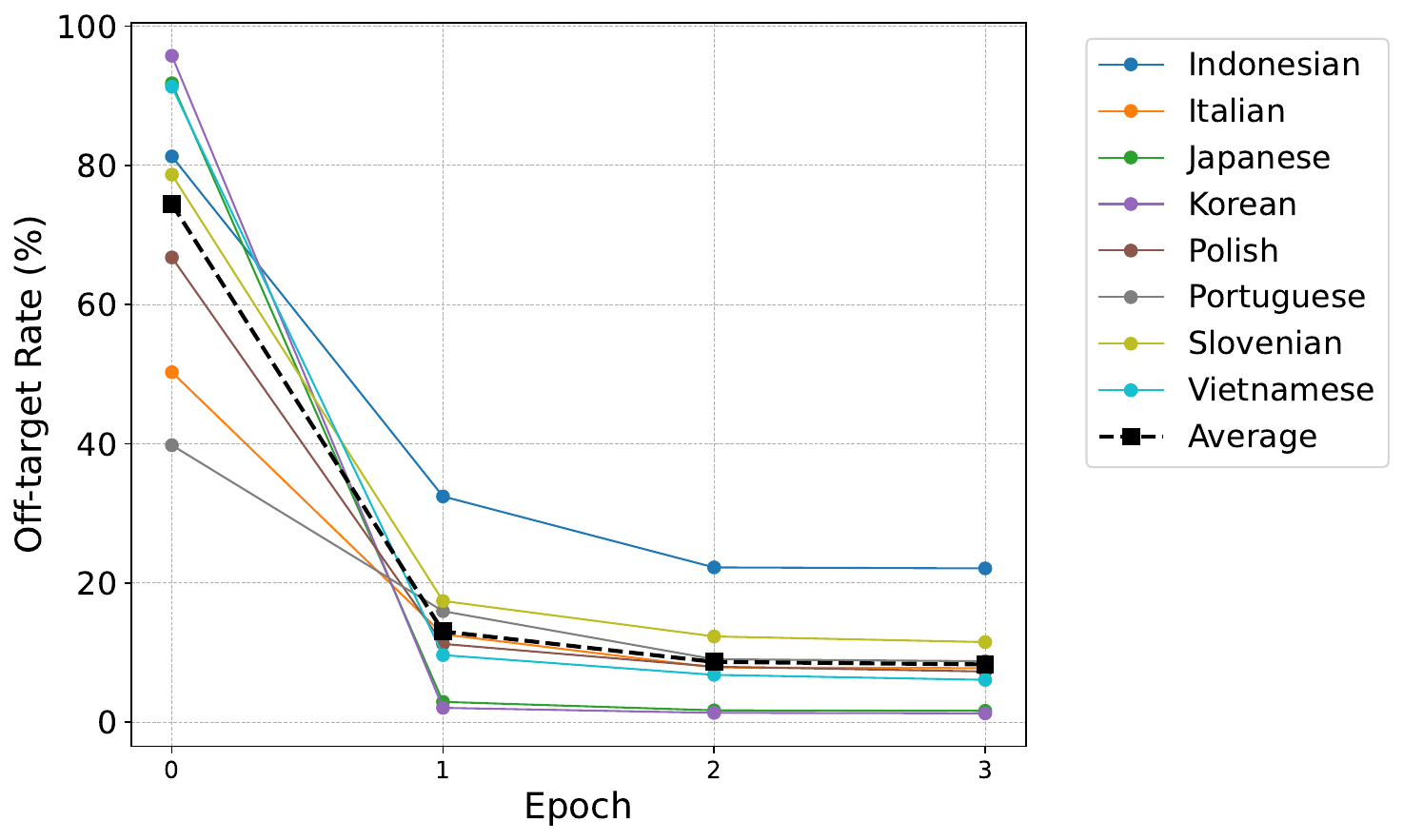}
    \caption{The occurrence rate of off-target issues in various languages during the SDRRL process.}
    \label{fig:offtar}
\end{figure}

We delve deeper into the effectiveness of SDRRL in alleviating off-target language issues during LLM responses. We evaluate the responses of LLaMA-2 on Dolly~\cite{DatabricksBlog2023DollyV2} and its multilingual extension, Bactrain-X~\cite{12li2023bactrianx}. As depicted in Figure~\ref{fig:offtar}, we showcase the variation in the off-target occurrence rates across each target language throughout the SDRRL process. This indicates that SDRRL plays a constructive role in mitigating off-target issues, ensuring consistency between the input and the response languages.

\section{Potential Risks of Our Method}
Because our method involves distilling knowledge from other target languages towards high-resource languages to achieve cross-linguistic alignment, it may lead to cultural unfairness for mid- and low-resource languages. For instance, after aligning to English using SDRRL, responses of LLMs in African languages may also adhere to the cultural practices and social norms of English.

\end{document}